\pdfoutput=1

\documentclass[11pt]{article}

\usepackage[]{acl}

\usepackage{times}
\usepackage{latexsym}

\usepackage[T1]{fontenc}

\usepackage[utf8]{inputenc}

\usepackage{microtype}

\usepackage{inconsolata}

\usepackage{graphicx}

\usepackage{colortbl}
\usepackage{xcolor}
\usepackage{multirow}
\usepackage{xspace}
\usepackage{tcolorbox}

\usepackage{algorithm}  
\usepackage{algorithmic} 
\usepackage[T1]{fontenc}
\usepackage[utf8]{inputenc}
\usepackage{latexsym}
\usepackage{amsmath}
\usepackage{amssymb}
\usepackage{bm}
\usepackage{bbm}
\usepackage{graphicx}
\usepackage{breakcites}
\usepackage{amsthm}
\usepackage{mdframed}
\usepackage{lipsum}
\usepackage[english]{babel}
\usepackage{delimset} %
\usepackage{tikz}
\usepackage{xcolor}
\usepackage{caption}
\usepackage{subcaption}
\usepackage{enumitem}

\usepackage{url}            %
\usepackage{booktabs}       %
\usepackage{amsfonts}       %
\usepackage{microtype}      %
\usepackage{wrapfig}
\usepackage{caption}

\usepackage{minitoc} %
\usepackage{xspace}
\usepackage{enumitem}
\usepackage{cleveref}

\usepackage{amsthm}
\usepackage{graphicx}
\usepackage{latexsym}
\usepackage{mathtools}
\usepackage{multirow}
\usepackage{fancyvrb}
\usepackage{adjustbox}

\crefname{section}{Section}{Sections}
\crefname{appendix}{Appendix}{Appendices}

\crefname{theorem}{Theorem}{Theorems}
\crefname{lemma}{Lemma}{Lemmas}
\crefname{corollary}{Corollary}{Corollaries}
\crefname{proposition}{Proposition}{Propositions}
\crefname{definition}{Definition}{Definitions}
\crefname{assumption}{Assumption}{Assumptions}

\Crefname{algorithm}{Algorithm}{Algorithms}
\crefname{figure}{Figure}{Figures}
\crefname{table}{Table}{Tables}

\usepackage{tasks}

\newtheorem{prompt}{Prompt}

\newcommand{\Sref}[1]{\S\ref{#1}}
\newcommand{\Fref}[1]{Figure~\ref{#1}}

\newcommand{\Aref}[1]{Appendix~\ref{#1}}

\newcommand{\Ourmethod}[1]{SelfReVision\xspace}
\newcommand{\ourmethod}[1]{SelfReVision\xspace}

\title{Making VLMs More Robot-Friendly: \\  Self-Critical Distillation of Low-Level Procedural Reasoning}

\author{
 \textbf{Chan Young Park\textsuperscript{1*}},
 \textbf{Jillian Fisher\textsuperscript{2*}},
 \textbf{Marius Memmel\textsuperscript{1}},
 \textbf{Dipika Khullar\textsuperscript{3}},
\\
 \textbf{Seoho Yun\textsuperscript{1}},
 \textbf{Abhishek Gupta\textsuperscript{1}},
 \textbf{Yejin Choi\textsuperscript{4}}
\\
\\
 \textsuperscript{1}Department of Computer Science, University of Washington,
 \textsuperscript{2}Department of Statistics, University of Washington, \\
 \textsuperscript{3}Independent Researcher,
 \textsuperscript{4}Department of Computer Science, Stanford University,\\
 \textsuperscript{*}Equal Contribution
\\
 \small{
   \textbf{Correspondence:} \href{mailto:email@domain}{chanpark@cs.washington.edu},
   \href{mailto:email@domain}{jrfish@uw.edu}
 }
}

\begin{document}
\maketitle
\begin{abstract}
Large language models (LLMs) have shown promise in robotic procedural planning, yet their human-centric reasoning often omits the low-level, grounded details needed for robotic execution. Vision-language models (VLMs) offer a path toward more perceptually grounded plans, but current methods either rely on expensive, large-scale models or are constrained to narrow simulation settings. We introduce \ourmethod{}\footnote{Code Available: \href{https://github.com/chan0park/SelfReVision}{https://github.com/chan0park/SelfReVision}}, a lightweight and scalable self-improvement framework for vision-language procedural planning. \ourmethod{} enables small VLMs to iteratively critique, revise, and verify their own plans—without external supervision or teacher models—drawing inspiration from chain-of-thought prompting and self-instruct paradigms. Through this self-distillation loop, models generate higher-quality, execution-ready plans that can be used both at inference and for continued fine-tuning. Using models varying from 3B to 72B, our results show that \ourmethod{} not only boosts performance over weak base VLMs but also outperforms models 100X the size, yielding improved control in downstream embodied tasks. 
\end{abstract} %
\vspace{-.5cm}
\section{Introduction}
\vspace{-.25cm}
\begin{figure}[t]
    \centering
    \includegraphics[width=0.95\linewidth]{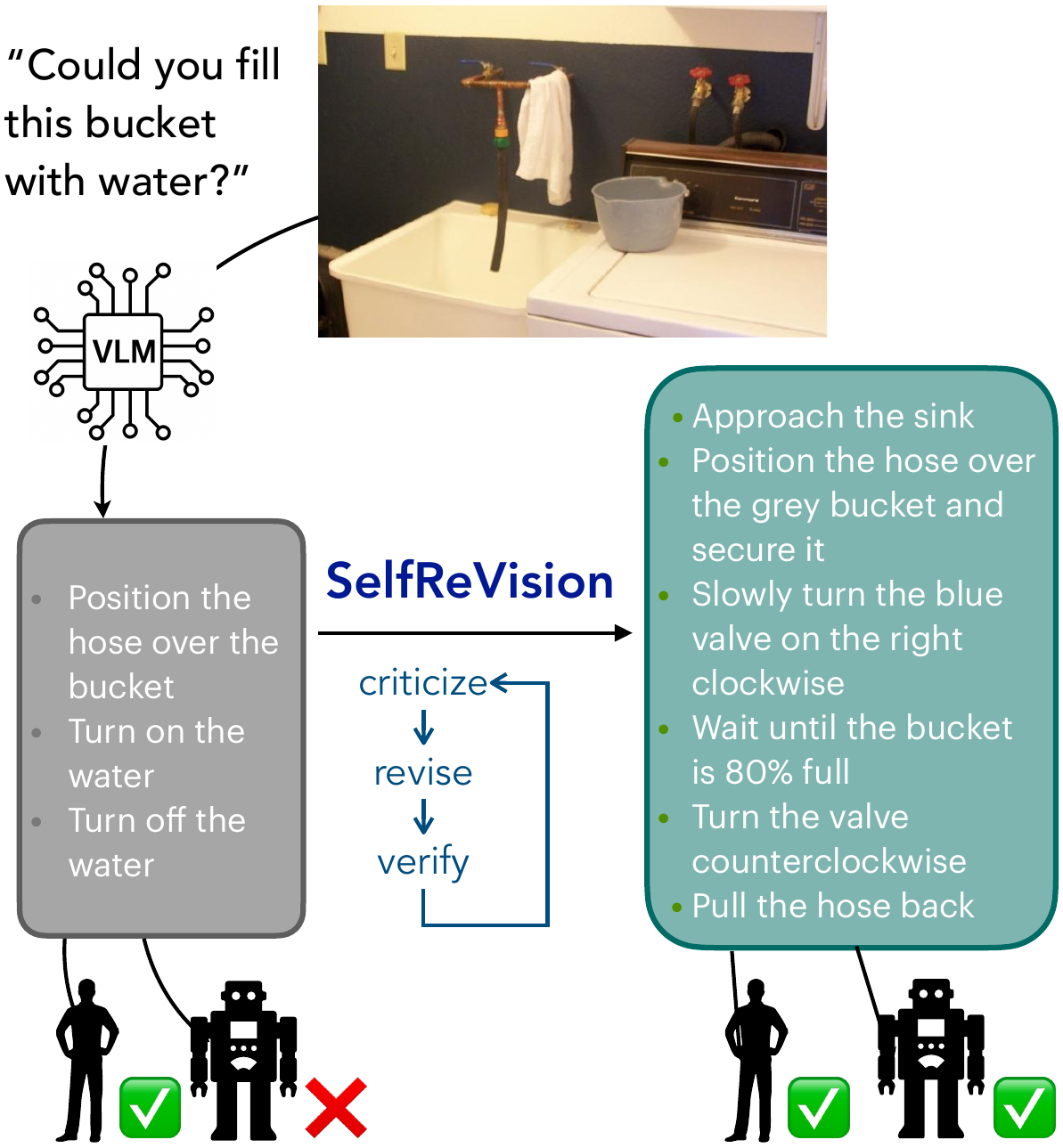}
    \caption{\textbf{Overview of \ourmethod{}}. VLMs tend to generate human-readable plans that are not detailed enough for robotic execution. \ourmethod, employs an iterative self-critique, revision, and verification process, to transform initial plans into actionable steps.}
    \label{fig:enter-label}
    \vspace{-20pt}
\end{figure}

Large language models (LLMs) have recently gained traction as a source of background knowledge for robotic applications, particularly in procedural planning tasks \cite{Huang2024UnderstandingTP, Ahn2022DoAI, Shi2025HiRO, Brahman2023PlaSmaMS}. Their broad pretraining and strong instruction-following capabilities make them appealing tools for generating step-by-step action sequences that, from a human perspective, appear sensible and coherent. Yet, a fundamental challenge remains: because LLMs are trained with human language and human preferences, they tend to generate plans in a way that is intuitive and meaningful to humans, rather than encoding the precise sensory or perceptual details that a robot would need to execute them. As a result, their plans often omit low-level, spatially grounded details essential for execution in the physical world. Consequently, when these plans are applied to robots, they may lead to uncertainty or mistakes in downstream tasks.

Bridging this gap calls for vision-language models (VLMs) that can reason over visual inputs to generate low-level procedural reasoning plans. Yet, current approaches face two critical shortcomings: they either (1) rely on overly specialized setups in simulation environments with limited real-world applicability \cite{Shi2025HiRO}, or (2) depend on massive, high-capacity models that are expensive to train and impractical to deploy in many real-world settings \cite{cheng2025, openai2024gpt4technicalreport, yang2024guidinglonghorizontaskmotion}. In contrast, many use cases, like in education, robotics, and resource-constrained environments, require solutions that are lightweight, data-efficient, and robust without relying on massive compute. \textit{We argue that strong vision planning can emerge even in smaller VLMs—if they are trained with the right inductive biases and self-improvement strategies.}

We present \ourmethod{}, a self-improvement framework for vision-language procedural planning based on iterative self-critiquing and self-refining. We show that this method enables small VLMs ranging as small as 3B to 72B, to enhance their performance through self-distillation, \textit{without any external supervision or teacher models}. Inspired by chain-of-thought reasoning and self-instruct methods, we break the task into a three-stage loop: the model first generates an initial plan from a prompt and image, then self-\textit{critiques} it with minimal guidance, self-\textit{revises} the plan accordingly, and finally selects the better of the two via a self-\textit{verification} step. This cycle repeats until the model produces a plan it deems better. The final plans, generated entirely by the model, can be used directly at inference or as self-supervised data to further fine-tune the model and reinforce improvements.

While self-critiquing and self-distillation has been explored in the LLM space \cite{NEURIPS2023_91edff07,gou2024critic}, its application to vision-language planning is largely underexplored. To our knowledge, \ourmethod{} is the first to adapt this paradigm for procedural planning with VLMs. Notably, we apply this method using small, weak base models to emphasize its potential as a tool for enhancing the capabilities of lightweight systems.  In addition, we provide a comprehensive ablation study of \ourmethod{}, showing insights into the role of each component and demonstrating why the iterative loop contributes to performance gains.

To rigorously assess our approach, we introduce a new vision-language evaluation dataset blending real-world and simulation-based visual procedural tasks—an underexplored combination in prior work. We demonstrate our improved VLM plans not only outperform their original base versions, but also surpass state-of-the-art VLMs of 100X larger size. Finally, we show that these enhanced procedural plans translate into better control and execution in downstream embodied agent tasks.

\vspace{-.25cm}
\section{Related Works}
\vspace{-.25cm}

\paragraph{Procedural Planning}
LLMs have become increasingly attractive for complex procedural planning tasks \cite{Huang2024UnderstandingTP}. Pretrained, off-the-shelf LLMs have shown strong performance in this area \cite{osti_10366294, Ahn2022DoAI}, and \citet{Brahman2023PlaSmaMS} further demonstrate that task-specific finetuning can boost their effectiveness even more. Beyond finetuning, another approach used to achieve procedural planning in LLMs is prompting pretrained models to interleave reasoning and action, improving adaptability and decision-making \cite{abs-2207-05608, osti_10451467}. Lastly, some methods instead aim to leverage LLMs for low-level action execution directly, bypassing high-level planning. For example, the Code-as-Policy framework prompts LLMs to produce structured, code-like plans that can be directly interpreted and executed as action sequences \cite{code_as_policy}.

Although LLMs have shown promising results in procedural planning, incorporating vision content can further broaden their practical utility and impact \cite{vla_survey, Lu2023MultimodalPP}. One approach to incorporating visual content is to adopt modular architectures, using specialized encoders to integrate multimodal information from different models \cite{Ilaslan2024VGTVPMP, Kalithasan2022LearningNP, cogact, song2023llmplanner,Yang2023MMREACTPC, zhu2023minigpt4enhancingvisionlanguageunderstanding}. Others enhance performance through finetuning \cite{driess2023palme, Shi2025HiRO} or by optimizing the prompts used with off-the-shelf models \cite{chen_et_al}. However, these systems are either large and resource-intensive \cite{driess2023palme} or rely on training data derived from even larger models \cite{Shi2025HiRO}. 
\vspace{-.05cm}

\paragraph{Self-Distillation and Self-Refinement}
With the advancement of vision and language models, research has explored using larger, more capable models to generate training data for fine-tuning smaller models, a process commonly referred to as knowledge distillation \cite{MOSLEMI2024100605,liu2023visualinstructiontuning, Xu2024ASO}. More recently, however, attention has turned toward self-distillation, in which a weaker model is used to improve itself without relying on a stronger teacher model.

A prominent form of self-distillation involves training data augmentation, where the model generates additional data to further fine-tune itself. This approach has yielded promising results across various domains, including instruction tuning \cite{wang-etal-2023-self-instruct}, preference modeling \cite{yang2024rlcdreinforcementlearningcontrastive}, and value alignment \cite{10.5555/3666122.3666237}. Beyond simply increasing the quantity of data, several studies have demonstrated that filtering the self-generated data can significantly enhance quality. Effective filtering strategies include promoting diversity \cite{wang-etal-2023-self-instruct}, selecting samples based on quality metrics \cite{jung-etal-2024-impossible}, and applying external scoring functions to encourage alignment with human values \cite{Gulcehre2023ReinforcedS}.

In addition to data generation and filtering, recent work has begun to explore ways in which models can analyze and guide themselves. For instance, some methods use interactive, chain-of-thought-style feedback to help weaker models arrive at correct answers for objective tasks such as math problems and question-answering tasks \cite{6b29035e8bde486ea9de38691dfb6e80, yu-etal-2024-teaching}. Similarly, \citet{10.5555/3666122.3668142} employed an LLM-as-Judge approach, using the weaker model itself as a reward model to learn stronger outputs on chatbot tasks. Lastly, similar to our approach, self-feedback and self-refinement techniques have shown promise for LLM tasks such as reasoning \cite{NEURIPS2023_81fde95c}, dialogue response \cite{NEURIPS2023_91edff07}, and mathematics \cite{gou2024critic, NEURIPS2023_91edff07}. However, these techniques have so far been primarily limited to objective tasks or use outside tools for critiquing \cite{gou2024critic, DBLP:journals/corr/abs-2411-16579}.

Self-distillation through self-refinement has been explored less frequently in the context of multi-modal models, but there are notable exceptions, particularly in image captioning. For example, \citet{wu2025sdrtenhancevisionlanguagemodels} proposed a method where the model generates intermediate reasoning hidden states, which are then used to retrain the base model, effectively improving performance through internal feedback. Other studies have leveraged self-distillation to augment human-annotated datasets, enriching the training corpus with additional synthetic examples \cite{NEURIPS2024_ed45d6a0, vila2}. These approaches suggest that even in multi-modal settings, self-distillation can provide valuable improvements when carefully designed.

\vspace{-.25cm}
\section{Methods}
\vspace{-.25cm}

\begin{algorithm}[t]
\caption{\ourmethod{}}
\label{alg:self-distillation}
\begin{algorithmic}
\REQUIRE Model $\theta$, Image $x$, Instruction $I$, 
\STATE Generate initial plan: $p_0 \gets \theta(x, I)$
\STATE Initialize: $p_{\text{curr}} \gets p_0$
\REPEAT
    \STATE Critique: $c \gets \text{Crit}(p_{\text{curr}})$ \hfill // Self-critique
    \STATE Revise: $p_{\text{rev}} \gets \text{Rev}(p_{\text{curr}}, c)$ \hfill // Generate improved plan
    \STATE Verify: $p_{\text{best}} \gets \text{Ver}(p_{\text{curr}}, p_{\text{rev}})$ \hfill // Choose better plan
    \IF{$p_{\text{best}} = p_{\text{rev}}$}
        \STATE \textbf{break} \hfill // Improvement found; terminate loop
    \ELSE
        \STATE \textbf{continue} \hfill // No improvement; keep current plan and revise again
    \ENDIF
\UNTIL{Convergence or max iterations reached}
\RETURN Final plan $p_{\text{curr}}$
    \vspace{-.1cm}
\end{algorithmic}
\end{algorithm}
 
\begin{table*}[t]
\centering
\resizebox{\textwidth}{!}{
\begin{tabular}{llccccccc|ccccccc}
\toprule
\multirow{2}{*}{} & \multirow{2}{*}{} & \multicolumn{7}{c|}{\textbf{Places}} & \multicolumn{7}{c}{\textbf{Simulation}} \\
& & Coverage & Ordering & Complete & Image. & Overall & \cellcolor{white}Imp.$\uparrow$ & +\#Inf & Coverage & Ordering & Complete & Image. & Overall & \cellcolor{white}Imp.$\uparrow$ & +\#Inf \\
\midrule
\multirow{5}{*}{\rotatebox{90}{Qwen-3B}} 
& $\Leftrightarrow$ GPT-4o & 1 $\Leftrightarrow$ 95 & 6 $\Leftrightarrow$ 83 & 1 $\Leftrightarrow$ 97 & 3 $\Leftrightarrow$ 60 & 0 $\Leftrightarrow$ 97 & \cellcolor{green!25}97 & 0 & 4 $\Leftrightarrow$ 95 & 7 $\Leftrightarrow$ 84 & 2 $\Leftrightarrow$ 98 & 5 $\Leftrightarrow$ 50 & 1 $\Leftrightarrow$ 98 & \cellcolor{green!25}97 & 0 \\
& $\Leftrightarrow$ PaliGemma & 91 $\Leftrightarrow$ 8 & 89 $\Leftrightarrow$ 5 & 90 $\Leftrightarrow$ 8 & 82 $\Leftrightarrow$ 4 & 92 $\Leftrightarrow$ 7 & \cellcolor{red!25}-85 & 0 & 95 $\Leftrightarrow$ 4 & 88 $\Leftrightarrow$ 7 & 93 $\Leftrightarrow$ 7 & 85 $\Leftrightarrow$ 3 & 95 $\Leftrightarrow$ 4 & \cellcolor{red!25}-91 & 0 \\
& $\Leftrightarrow$ Best-of-N & 31 $\Leftrightarrow$ 44 & 32 $\Leftrightarrow$ 41 & 41 $\Leftrightarrow$ 54 & 18 $\Leftrightarrow$ 28 & 38 $\Leftrightarrow$ 58 & \cellcolor{green!10}20 & 6 & 46 $\Leftrightarrow$ 35 & 43 $\Leftrightarrow$ 23 & 53 $\Leftrightarrow$ 40 & 16 $\Leftrightarrow$ 21 & 51 $\Leftrightarrow$ 40 & \cellcolor{red!10}-11 & 6 \\
& $\Leftrightarrow$ SelfReVision & 9 $\Leftrightarrow$ 52 & 17 $\Leftrightarrow$ 28 & 6 $\Leftrightarrow$ 62 & 9 $\Leftrightarrow$ 12 & 15 $\Leftrightarrow$ 61 & \cellcolor{green!25}46 & 9.1 & 6 $\Leftrightarrow$ 52 & 20 $\Leftrightarrow$ 26 & 6 $\Leftrightarrow$ 74 & 11 $\Leftrightarrow$ 18 & 12 $\Leftrightarrow$ 67 & \cellcolor{green!25}55 & 7.2 \\
& $\Leftrightarrow$ SelfReVision+SFT &  25 $\Leftrightarrow$ 59 & 25 $\Leftrightarrow$ 49  & 29 $\Leftrightarrow$ 68  & 25 $\Leftrightarrow$ 26 & 30 $\Leftrightarrow$ 68 & \cellcolor{green!15} 38 & 0 & 29 $\Leftrightarrow$ 58 & 30 $\Leftrightarrow$ 37 & 34 $\Leftrightarrow$ 63 &  17 $\Leftrightarrow$ 25 & 32 $\Leftrightarrow$ 61 & \cellcolor{green!15} 29 & 0 \\
\midrule

\multirow{5}{*}{\rotatebox{90}{Gemma-4B}} 
& $\Leftrightarrow$ GPT-4o & 8 $\Leftrightarrow$ 80 & 15 $\Leftrightarrow$ 69 & 12 $\Leftrightarrow$ 81 & 13 $\Leftrightarrow$ 54 & 11 $\Leftrightarrow$ 89 & \cellcolor{green!25}78 & 0 & 15 $\Leftrightarrow$ 59 & 28 $\Leftrightarrow$ 53 & 31 $\Leftrightarrow$ 68 & 22 $\Leftrightarrow$ 35 & 25 $\Leftrightarrow$ 73 & \cellcolor{green!10}48 & 0 \\
& $\Leftrightarrow$ PaliGemma & 97 $\Leftrightarrow$ 3 & 92 $\Leftrightarrow$ 4 & 100 $\Leftrightarrow$ 0 & 87 $\Leftrightarrow$ 4 & 97 $\Leftrightarrow$ 3 & \cellcolor{red!25}-94 & 0 & 98 $\Leftrightarrow$ 2 & 95 $\Leftrightarrow$ 3 & 98 $\Leftrightarrow$ 1 & 91 $\Leftrightarrow$ 1 & 98 $\Leftrightarrow$ 2 & \cellcolor{red!25}-96 & 0 \\
& $\Leftrightarrow$ Best-of-N & 26 $\Leftrightarrow$ 44 & 30 $\Leftrightarrow$ 35 & 30 $\Leftrightarrow$ 54 & 18 $\Leftrightarrow$ 32 & 33 $\Leftrightarrow$ 57 & \cellcolor{green!10}24 & 6 & 28 $\Leftrightarrow$ 39 & 29 $\Leftrightarrow$ 30 & 40 $\Leftrightarrow$ 51 & 14 $\Leftrightarrow$ 27 & 39 $\Leftrightarrow$ 52 & \cellcolor{green!10}13 & 6 \\
& $\Leftrightarrow$ SelfReVision & 8 $\Leftrightarrow$ 73 & 26 $\Leftrightarrow$ 53 & 6 $\Leftrightarrow$ 88 & 16 $\Leftrightarrow$ 42 & 8 $\Leftrightarrow$ 86 & \cellcolor{green!25}78 & 8.9 & 10 $\Leftrightarrow$ 73 & 33 $\Leftrightarrow$ 53 & 5 $\Leftrightarrow$ 90 & 17 $\Leftrightarrow$ 25 & 13 $\Leftrightarrow$ 82 & \cellcolor{green!25}69 & 7.0 \\
& $\Leftrightarrow$ SelfReVision+SFT & 32 $\Leftrightarrow$ 56 & 33 $\Leftrightarrow$ 57 & 39 $\Leftrightarrow$ 58 & 27 $\Leftrightarrow$ 41 & 36 $\Leftrightarrow$ 60 & \cellcolor{green!10}24 & 0 & 33 $\Leftrightarrow$ 45&35 $\Leftrightarrow$ 47 &38 $\Leftrightarrow$ 56 &28 $\Leftrightarrow$ 45 &36 $\Leftrightarrow$ 54 &\cellcolor{green!10}  11 &0\\
\midrule
\multirow{5}{*}{\rotatebox{90}{Qwen-7B}} 
& $\Leftrightarrow$ GPT-4o & 11 $\Leftrightarrow$ 76 & 19 $\Leftrightarrow$ 66 & 11 $\Leftrightarrow$ 85 & 10 $\Leftrightarrow$ 48 & 10 $\Leftrightarrow$ 86 & \cellcolor{green!25}76 & 0 & 11 $\Leftrightarrow$ 76 & 18 $\Leftrightarrow$ 66 & 10 $\Leftrightarrow$ 87 & 10 $\Leftrightarrow$ 44 & 10 $\Leftrightarrow$ 86 & \cellcolor{green!25}76 & 0 \\
& $\Leftrightarrow$ PaliGemma & 99 $\Leftrightarrow$ 0 & 95 $\Leftrightarrow$ 2 & 100 $\Leftrightarrow$ 0 & 82 $\Leftrightarrow$ 3 & 99 $\Leftrightarrow$ 1 & \cellcolor{red!25}-98 & 0 & 98 $\Leftrightarrow$ 2 & 96 $\Leftrightarrow$ 1 & 99 $\Leftrightarrow$ 1 & 87 $\Leftrightarrow$ 1 & 98 $\Leftrightarrow$ 2 & \cellcolor{red!25}-96 & 0 \\
& $\Leftrightarrow$ Best-of-N & 29 $\Leftrightarrow$ 48 & 31 $\Leftrightarrow$ 43 & 42 $\Leftrightarrow$ 51 & 12 $\Leftrightarrow$ 36 & 38 $\Leftrightarrow$ 58 & \cellcolor{green!10}20 & 6 & 35 $\Leftrightarrow$ 41 & 32 $\Leftrightarrow$ 25 & 43 $\Leftrightarrow$ 47 & 15 $\Leftrightarrow$ 25 & 41 $\Leftrightarrow$ 49 & \cellcolor{green!10}8 & 6 \\
& $\Leftrightarrow$ SelfReVision & 3 $\Leftrightarrow$ 71 & 30 $\Leftrightarrow$ 31 & 3 $\Leftrightarrow$ 91 & 17 $\Leftrightarrow$ 38 & 9 $\Leftrightarrow$ 82 & \cellcolor{green!25}73 & 9.2 & 2 $\Leftrightarrow$ 75 & 38 $\Leftrightarrow$ 21 & 5 $\Leftrightarrow$ 89 & 12 $\Leftrightarrow$ 16 & 7 $\Leftrightarrow$ 86 & \cellcolor{green!25}79 & 10.0 \\
& $\Leftrightarrow$ SelfReVision+SFT & 20 $\Leftrightarrow$ 64 & 35 $\Leftrightarrow$ 46 & 20 $\Leftrightarrow$ 77 & 22 $\Leftrightarrow$ 43 & 18 $\Leftrightarrow$ 75 & \cellcolor{green!10}57 & 0 & 17 $\Leftrightarrow$ 75&37 $\Leftrightarrow$ 44 & 19 $\Leftrightarrow$ 79 & 19 $\Leftrightarrow$ 22 & 21 $\Leftrightarrow$ 75 & \cellcolor{green!15}54 & 0\\
\midrule
\multirow{5}{*}{\rotatebox{90}{Gemma-12B}} 
& $\Leftrightarrow$ GPT-4o & 24 $\Leftrightarrow$ 49 & 22 $\Leftrightarrow$ 54 & 37 $\Leftrightarrow$ 56 & 25 $\Leftrightarrow$ 31 & 32 $\Leftrightarrow$ 56 & \cellcolor{green!10}24 & 0 & 38 $\Leftrightarrow$ 41 & 32 $\Leftrightarrow$ 52 & 44 $\Leftrightarrow$ 54 & 22 $\Leftrightarrow$ 32 & 44 $\Leftrightarrow$ 55 & \cellcolor{green!10}11 & 0 \\
& $\Leftrightarrow$ PaliGemma & 100 $\Leftrightarrow$ 0 & 97 $\Leftrightarrow$ 1 & 100 $\Leftrightarrow$ 0 & 90 $\Leftrightarrow$ 0 & 100 $\Leftrightarrow$ 0 & \cellcolor{red!25}-100 & 0 & 100 $\Leftrightarrow$ 0 & 100 $\Leftrightarrow$ 0 & 100 $\Leftrightarrow$ 0 & 91 $\Leftrightarrow$ 1 & 100 $\Leftrightarrow$ 0 & \cellcolor{red!25}-100 & 0 \\
& $\Leftrightarrow$ Best-of-N & 23 $\Leftrightarrow$ 40 & 31 $\Leftrightarrow$ 33 & 33 $\Leftrightarrow$ 55 & 19 $\Leftrightarrow$ 33 & 29 $\Leftrightarrow$ 54 & \cellcolor{green!10}25 & 6 & 18 $\Leftrightarrow$ 41 & 21 $\Leftrightarrow$ 46 & 29 $\Leftrightarrow$ 55 & 8 $\Leftrightarrow$ 26 & 28 $\Leftrightarrow$ 61 & \cellcolor{green!10}33 & 6 \\
& $\Leftrightarrow$ SelfReVision & 8 $\Leftrightarrow$ 79 & 51 $\Leftrightarrow$ 31 & 6 $\Leftrightarrow$ 91 & 36 $\Leftrightarrow$ 35 & 10 $\Leftrightarrow$ 80 & \cellcolor{green!25}70 & 6.7 & 8 $\Leftrightarrow$ 84 & 50 $\Leftrightarrow$ 35 & 6 $\Leftrightarrow$ 93 & 29 $\Leftrightarrow$ 19 & 11 $\Leftrightarrow$ 81 & \cellcolor{green!25}70 & 6.6 \\
& $\Leftrightarrow$ SelfReVision+SFT & 24 $\Leftrightarrow$ 64& 43 $\Leftrightarrow$ 45 & 22 $\Leftrightarrow$  77 & 38 $\Leftrightarrow$ 26 & 23 $\Leftrightarrow$ 72& \cellcolor{green!15}49 & 0 & 16 $\Leftrightarrow$ 71 & 49 $\Leftrightarrow$ 32& 17 $\Leftrightarrow$ 81& 32 $\Leftrightarrow$ 22 & 24 $\Leftrightarrow$ 70 & \cellcolor{green!15} 46 & 0\\

\midrule
\multirow{4}{*}{\rotatebox{90}{Gemma-27B}} 
& $\Leftrightarrow$ GPT-4o & 31 $\Leftrightarrow$ 45 & 29 $\Leftrightarrow$ 50 & 42 $\Leftrightarrow$ 53 & 31 $\Leftrightarrow$ 31 & 39 $\Leftrightarrow$ 53 & \cellcolor{green!10}14 & 0 & 38 $\Leftrightarrow$ 34 & 30 $\Leftrightarrow$ 46 & 46 $\Leftrightarrow$ 47 & 34 $\Leftrightarrow$ 17 & 46 $\Leftrightarrow$ 48 & \cellcolor{green!10}2 & 0 \\
& $\Leftrightarrow$ PaliGemma & 100 $\Leftrightarrow$ 0 & 98 $\Leftrightarrow$ 2 & 100 $\Leftrightarrow$ 0 & 98 $\Leftrightarrow$ 2 & 100 $\Leftrightarrow$ 0 & \cellcolor{red!25}-100 & 0 & 99 $\Leftrightarrow$ 1 & 96 $\Leftrightarrow$ 2 & 99 $\Leftrightarrow$ 1 & 95 $\Leftrightarrow$ 1 & 99 $\Leftrightarrow$ 1 & \cellcolor{red!25}-98 & 0 \\
& $\Leftrightarrow$ Best-of-N & 22 $\Leftrightarrow$ 37 & 26 $\Leftrightarrow$ 28 & 34 $\Leftrightarrow$ 55 & 17 $\Leftrightarrow$ 24 & 35 $\Leftrightarrow$ 53 & \cellcolor{green!10}18 & 6 & 23 $\Leftrightarrow$ 38 & 27 $\Leftrightarrow$ 28 & 37 $\Leftrightarrow$ 54 & 15 $\Leftrightarrow$ 26 & 31 $\Leftrightarrow$ 58 & \cellcolor{green!10}27 & 6 \\
& $\Leftrightarrow$ SelfReVision & 6 $\Leftrightarrow$ 85 & 50 $\Leftrightarrow$ 34 & 1 $\Leftrightarrow$ 97 & 28 $\Leftrightarrow$ 21 & 7 $\Leftrightarrow$ 86 & \cellcolor{green!25}79 & 6.6 & 4 $\Leftrightarrow$ 89 & 39 $\Leftrightarrow$ 48 & 3 $\Leftrightarrow$ 97 & 36 $\Leftrightarrow$ 22 & 7 $\Leftrightarrow$ 88 & \cellcolor{green!25}81 & 6.2 \\
\\[-10pt]
\midrule
\multirow{4}{*}{\rotatebox{90}{Qwen-32B}} 
& $\Leftrightarrow$ GPT-4o & 51 $\Leftrightarrow$ 20 & 28 $\Leftrightarrow$ 40 & 74 $\Leftrightarrow$ 20 & 26 $\Leftrightarrow$ 29 & 63 $\Leftrightarrow$ 32 & \cellcolor{red!25}-31 & 0 & 52 $\Leftrightarrow$ 19 & 31 $\Leftrightarrow$ 44 & 74 $\Leftrightarrow$ 17 & 25 $\Leftrightarrow$ 27 & 71 $\Leftrightarrow$ 26 & \cellcolor{red!25}-45 & 0 \\
& $\Leftrightarrow$ PaliGemma & 100 $\Leftrightarrow$ 0 & 99 $\Leftrightarrow$ 0 & 100 $\Leftrightarrow$ 0 & 90 $\Leftrightarrow$ 2 & 100 $\Leftrightarrow$ 0 & \cellcolor{red!25}-100 & 0 & 100 $\Leftrightarrow$ 0 & 98 $\Leftrightarrow$ 1 & 100 $\Leftrightarrow$ 0 & 90 $\Leftrightarrow$ 0 & 100 $\Leftrightarrow$ 0 & \cellcolor{red!25}-100 & 0 \\
& $\Leftrightarrow$ Best-of-N & 22 $\Leftrightarrow$ 43 & 27 $\Leftrightarrow$ 29 & 31 $\Leftrightarrow$ 65 & 21 $\Leftrightarrow$ 25 & 37 $\Leftrightarrow$ 54 & \cellcolor{green!10}17 & 6 & 18 $\Leftrightarrow$ 52 & 21 $\Leftrightarrow$ 41 & 24 $\Leftrightarrow$ 65 & 17 $\Leftrightarrow$ 17 & 27 $\Leftrightarrow$ 65 & \cellcolor{green!10}38 & 6 \\
& $\Leftrightarrow$ SelfReVision & 2 $\Leftrightarrow$ 60 & 28 $\Leftrightarrow$ 23 & 1 $\Leftrightarrow$ 62 & 21 $\Leftrightarrow$ 9 & 3 $\Leftrightarrow$ 56 & \cellcolor{green!25}53 & 16 & 0 $\Leftrightarrow$ 69 & 28 $\Leftrightarrow$ 31 & 0 $\Leftrightarrow$ 78 & 15 $\Leftrightarrow$ 14 & 4 $\Leftrightarrow$ 72 & \cellcolor{green!25}68 & 12.7 \\
\midrule
\multirow{4}{*}{\rotatebox{90}{Qwen-72B}} 
& $\Leftrightarrow$ GPT-4o & 14 $\Leftrightarrow$ 61 & 20 $\Leftrightarrow$ 59 & 17 $\Leftrightarrow$ 76 & 17 $\Leftrightarrow$ 35 & 15 $\Leftrightarrow$ 76 & \cellcolor{green!25}61 & 0 & 11 $\Leftrightarrow$ 62 & 15 $\Leftrightarrow$ 70 & 19 $\Leftrightarrow$ 79 & 16 $\Leftrightarrow$ 43 & 13 $\Leftrightarrow$ 82 & \cellcolor{green!25}69 & 0 \\
& $\Leftrightarrow$ PaliGemma & 98 $\Leftrightarrow$ 1 & 95 $\Leftrightarrow$ 4 & 100 $\Leftrightarrow$ 0 & 88 $\Leftrightarrow$ 4 & 98 $\Leftrightarrow$ 1 & \cellcolor{green!25}97 & 0 & 99 $\Leftrightarrow$ 1 & 98 $\Leftrightarrow$ 1 & 100 $\Leftrightarrow$ 0 & 94 $\Leftrightarrow$ 0 & 99 $\Leftrightarrow$ 1 & \cellcolor{red!25}-98 & 0\\
& $\Leftrightarrow$ Best-of-N & 23 $\Leftrightarrow$ 56 & 23 $\Leftrightarrow$ 52 & 25 $\Leftrightarrow$ 72 & 15 $\Leftrightarrow$ 37 & 25 $\Leftrightarrow$ 72 & \cellcolor{green!10}47 & 6 & 12 $\Leftrightarrow$ 60 & 21 $\Leftrightarrow$ 40 & 11 $\Leftrightarrow$ 85 & 14 $\Leftrightarrow$ 27 & 13 $\Leftrightarrow$ 81 & \cellcolor{green!25}68 & 6 \\
& $\Leftrightarrow$ SelfReVision & 4 $\Leftrightarrow$ 89 & 49 $\Leftrightarrow$ 39 & 2 $\Leftrightarrow$ 98 & 21 $\Leftrightarrow$ 38 & 6 $\Leftrightarrow$ 85 & \cellcolor{green!25}79 & 7.3 & 5 $\Leftrightarrow$ 90 & 31 $\Leftrightarrow$ 53 & 2 $\Leftrightarrow$ 97 & 25 $\Leftrightarrow$ 26 & 8 $\Leftrightarrow$ 89 & \cellcolor{green!25}81 & 6.5 \\

\bottomrule
\end{tabular}
}
\vspace{-.25cm}
\caption{\textbf{Win rate comparison of baseline models and \ourmethod{} against initial plan $p_0$}, across two datasets (\textsc{Places} and \textsc{Simulation}). Evaluation is done using GPT\-4o as judge across five dimensions, including overall improvement (Imp.) and number of inference calls (+\#Inf). Higher improvement indicates better plan quality.}
    \vspace{-.5cm}
\label{tab:full-results}
\end{table*}
 \textit{Procedural planning} involves generating a step-by-step plan to achieve a goal. We focus on open-ended, multi-step tasks with diverse, valid solutions. Unlike prior work relying on powerful LLMs in purely textual settings, we tackle a harder, more realistic problem: vision-grounded procedural planning using only weak VLMs. This multimodal setup adds complexity—plans must align with user intent and visual constraints like spatial layout, semantics, and object presence. We further restrict ourselves to low-capacity models, reflecting deployment in resource-limited settings without large teacher models or gold labels. To meet this challenge, we propose a self-distillation framework where a weak model improves through its own reasoning, via a structured loop of critique, revision, and verification, without external supervision or extra data.

\vspace{-.05cm}
\paragraph{Self-Distillation via Self-Improvement}
We build on the principle of self-distillation, a training paradigm where a model improves itself by learning from its own outputs. Unlike classical knowledge distillation, which requires a stronger teacher model, our approach is entirely self-supervised. Let $\theta$ denote a base model. We define a self-distilled dataset $D$ as:
\vspace{-0.7em}
\begin{align*}
D = (x, y, \phi_{\text{sd}}(x, y)) \mid x \sim \mathcal{X},\ y \sim p_\theta(y \mid x, I),
\end{align*}
\vspace{-22pt}

where $x$ is an input prompt, $I$ is an instruction or task description, and $y$ is the model’s own initial plan output. The transformation function $\phi_{\text{sd}}$ refines this output via a structured process involving targeted critique and revision.

\vspace{-.05cm}
\paragraph{\ourmethod{}}
We introduce \ourmethod{}, a three-stage Criticize–Revise–Verify pipeline to instantiate $\phi_{\text{sd}}$. This process encourages the model to iteratively refine its outputs via structured introspection:
\begin{itemize}[itemsep=0pt, topsep=5pt,leftmargin=*]
    
\item Criticize ($\text{Crit}$): The model generates an initial plan $p_0 = \theta(x,I)$, which may be vague, image-agnostic, or incomplete. We then prompt the model to produce a critical self-assessment $\text{Crit}(p_0)$.

\item Revise ($\text{Rev}$): Using its self-generated critique, the model produces a revised plan $p_1 = \text{Rev}(p_0, \text{Crit}(p_0))$. This phase encourages localized, meaningful improvements, splitting complex revisions into manageable subgoals via chain-of-thought prompting.

\item Verify ($\text{Ver}$): Finally, the model evaluates both $p_0$ and $p_1$ to decide which is superior: $p_{\text{best}} = \text{Ver}(p_0, p_1)$. If the revised plan is preferred, the process terminates, if is not preferred then the process continues recursively until a better plan is produced.
\end{itemize}

The iterative nature of this loop is formalized in \cref{alg:self-distillation}, and can be used to run for any threshold amount of refinement loops (i.e. rounds). It can also be used to generate a set amount of final plans $p_{curr}$ that can then be compared to the baseline or each-other. This closed-loop formulation mimics aspects of human self-improvement, which identifies flaws, attempt revision, and critically evaluate the result. 

\vspace{-.05cm}
\paragraph{Inference vs. Finetuning} \ourmethod{} generates curated outputs through self-distillation, which can be leveraged in two ways: used directly at inference time or as training data for finetuning. Using \ourmethod{} at inference time requires no model updates and allows fast deployment, but may incur computational overhead or complexity in orchestration. In contrast, finetuning incorporates the improvements directly into the model, enabling faster inference and better generalization, but requires additional training time and resources. The choice depends on the desired balance between flexibility, performance, and scalability.

\vspace{-.25cm}
\section{Experiments}
\vspace{-.25cm}
\begin{figure}
    \centering
    \includegraphics[width=1\linewidth]{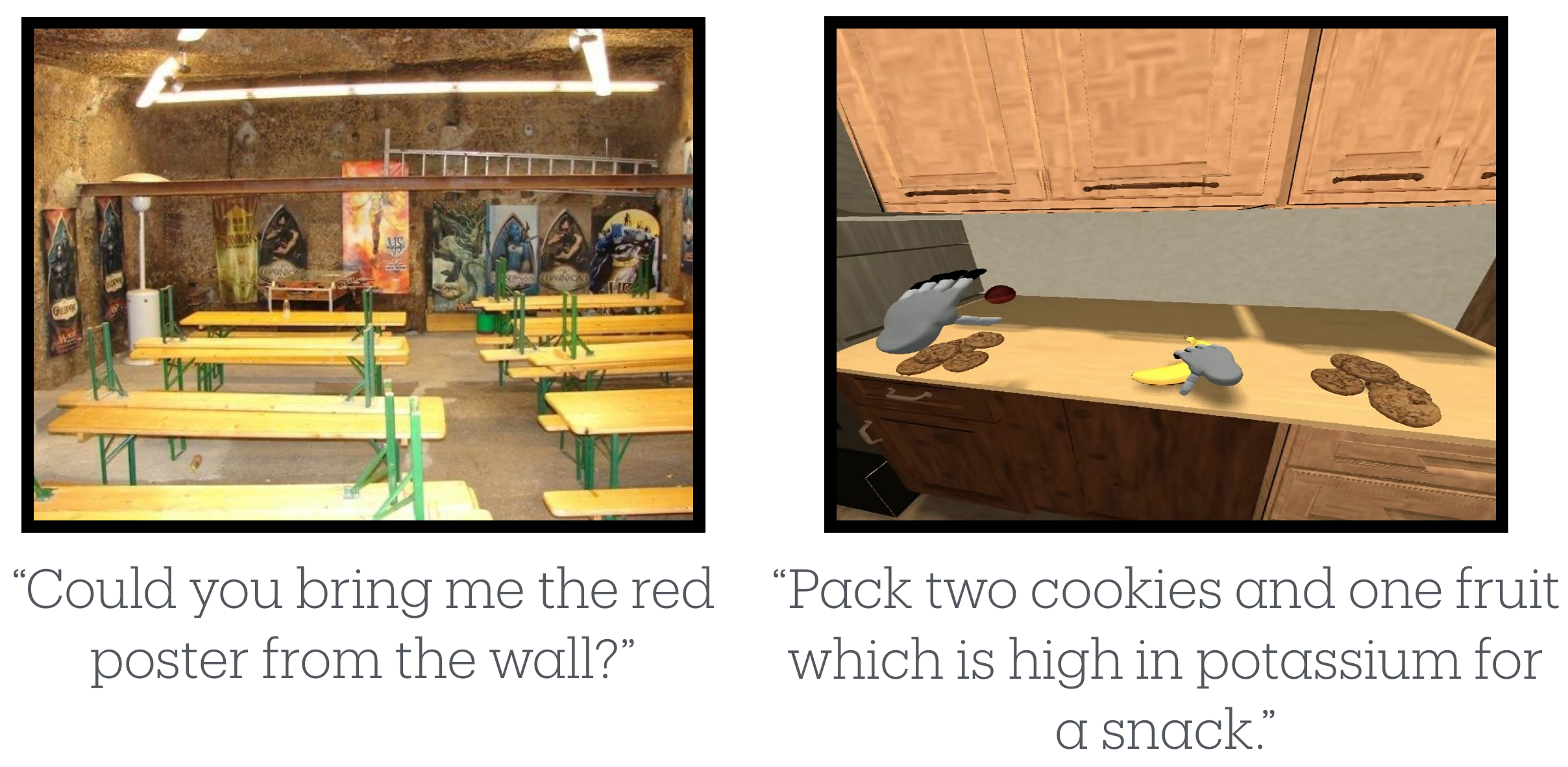}
    \vspace{-.5cm}
    \caption{\textbf{Evaluation examples} from the real-world \textsc{Places} dataset \cite{zhou2017places} (right) and from the \textsc{Simulation} dataset, VirtualHome \cite{puig2018virtualhome} and BEHAVIOR-100 \cite{pmlr-v164-srivastava22a} (left).}
    \label{fig:data_example}
    \vspace{-.5cm}
\end{figure}
\begin{table*}[t]
\centering
\resizebox{\textwidth}{!}{
\begin{tabular}{llcccccc|cccccc}
\toprule
\multirow{2}{*}{}  & \multicolumn{7}{c|}{\textbf{Places}} & \multicolumn{6}{c}{\textbf{Simulation}} \\
& & Coverage & Ordering & Complete & Image. & Overall & \cellcolor{white}Imp.$\uparrow$  & Coverage & Ordering & Complete & Image. & Overall & \cellcolor{white}Imp.$\uparrow$\\
\midrule
\multirow{7}{*}{\rotatebox{90}{GPT-4o}} 
& $\Leftrightarrow$ Qwen-3B & 91 $\Leftrightarrow$ 7 & 82 $\Leftrightarrow$ 9 & 91 $\Leftrightarrow$ 7 & 56 $\Leftrightarrow$ 6 & 92 $\Leftrightarrow$ 3 & \cellcolor{red!25}-89 & 83 $\Leftrightarrow$ 9 & 82 $\Leftrightarrow$ 5 & 87 $\Leftrightarrow$ 13 & 52 $\Leftrightarrow$ 9 & 93 $\Leftrightarrow$ 3 & \cellcolor{red!25}-90 \\
& $\Leftrightarrow$ Gemma-4B & 47 $\Leftrightarrow$ 32 & 73 $\Leftrightarrow$ 17 & 56 $\Leftrightarrow$ 43 & 48 $\Leftrightarrow$ 18 & 57 $\Leftrightarrow$ 36 & \cellcolor{red!20}-21 & 44 $\Leftrightarrow$37 & 71 $\Leftrightarrow$23 & 47 $\Leftrightarrow$ 48 & 45 $\Leftrightarrow$ 14 & 58 $\Leftrightarrow$ 38 & \cellcolor{red!20}-20 \\
& $\Leftrightarrow$ Qwen-7B & 54 $\Leftrightarrow$ 27 & 69 $\Leftrightarrow$ 21 & 55 $\Leftrightarrow$ 42 & 44 $\Leftrightarrow$ 15 & 60 $\Leftrightarrow$ 30 & \cellcolor{red!20}-30 & 55 $\Leftrightarrow$ 30 & 75 $\Leftrightarrow$ 15 & 61 $\Leftrightarrow$ 37 & 42 $\Leftrightarrow$ 15 & 62 $\Leftrightarrow$ 31 & \cellcolor{red!20} -31 \\
& $\Leftrightarrow$ Gemma-12B & 23 $\Leftrightarrow$ 67 & 69 $\Leftrightarrow$ 21 & 16 $\Leftrightarrow$ 82 & 45 $\Leftrightarrow$ 25 & 26 $\Leftrightarrow$ 65 & \cellcolor{green!20}39  & 17 $\Leftrightarrow$ 74 & 73 $\Leftrightarrow$ 22 & 15 $\Leftrightarrow$ 84 & 39 $\Leftrightarrow$ 20 & 25 $\Leftrightarrow$ 68 & \cellcolor{green!15}43 \\
& $\Leftrightarrow$ Gemma-27B & 15 $\Leftrightarrow$ 73 & 46 $\Leftrightarrow$ 40 & 11 $\Leftrightarrow$ 88 & 38 $\Leftrightarrow$ 24 & 20 $\Leftrightarrow$ 70 & \cellcolor{green!20}50  & 10 $\Leftrightarrow$ 82 & 41 $\Leftrightarrow$ 42 & 7 $\Leftrightarrow$ 92 & 34 $\Leftrightarrow$ 19 & 9 $\Leftrightarrow$ 81 & \cellcolor{green!25}72 \\
& $\Leftrightarrow$ Qwen-32B & 11 $\Leftrightarrow$ 76 & 53 $\Leftrightarrow$ 32 & 9 $\Leftrightarrow$ 91 &34 $\Leftrightarrow$ 20 & 15 $\Leftrightarrow$ 82 & \cellcolor{green!20}67  & 5 $\Leftrightarrow$ 83 & 44 $\Leftrightarrow$ 36 & 7 $\Leftrightarrow$ 91 & 38 $\Leftrightarrow$ 12 & 10 $\Leftrightarrow$ 84 & \cellcolor{green!25}74 \\
& $\Leftrightarrow$ Qwen-72B &  27 $\Leftrightarrow$ 63 & 72 $\Leftrightarrow$ 18  & 19 $\Leftrightarrow$ 78  & 42 $\Leftrightarrow$ 33 & 32 $\Leftrightarrow$ 58 & \cellcolor{green!15} 26 & 19 $\Leftrightarrow$ 71 & 65 $\Leftrightarrow$ 25 & 19 $\Leftrightarrow$ 81 &  32 $\Leftrightarrow$ 17 & 25 $\Leftrightarrow$ 69 & \cellcolor{green!15} 44 \\
\bottomrule
\end{tabular}}
\vspace{-.25cm}
\caption{\textbf{Win rate comparison of GPT\-4o and \ourmethod{} plans directly}, across two datasets (\textsc{Places} and \textsc{Simulation}). Evaluation is done using GPT\-4o as judge across five dimensions, including overall improvement (Imp.). Higher improvement indicates better plan quality.}
    \vspace{-.5cm}
\label{tab:gpt4o}
\end{table*}
 
\begin{table}[t]
\centering
\resizebox{\columnwidth}{!}{
\begin{tabular}{llccccc|c}
\toprule
 &  & Coverage & Ordering & Complete & Image & Overall & Imp.$\uparrow$ \\
\midrule
\multirow{4}{*}{\rotatebox{90}{Places}} 
& CRV & 5.7 $\Leftrightarrow$ 72.7 & 35.9 $\Leftrightarrow$ 34.1 & 3.6 $\Leftrightarrow$ 84.1 & 21.1 $\Leftrightarrow$ 27.9 & 8.3 $\Leftrightarrow$ 76.6 & \textbf{68.3} \\
& CR  & 9.4 $\Leftrightarrow$ 67.0 & 37.4 $\Leftrightarrow$ 30.7 & 7.1 $\Leftrightarrow$ 78.1 & 25.6 $\Leftrightarrow$ 24.4 & 11.3 $\Leftrightarrow$ 70.3 & \textbf{59.0} \\
& RV  & 7.4 $\Leftrightarrow$ 34.7 & 14.4 $\Leftrightarrow$ 19.4 & 7.1 $\Leftrightarrow$ 56.6 & 6.6 $\Leftrightarrow$ 26.4 & 13.3 $\Leftrightarrow$ 60.0 & \textbf{46.7} \\
& R   & 9.6 $\Leftrightarrow$ 32.7 & 12.9 $\Leftrightarrow$ 18.4 & 10.3 $\Leftrightarrow$ 52.1 & 8.0 $\Leftrightarrow$ 24.0 & 16.9 $\Leftrightarrow$ 55.0 & \textbf{38.1} \\
\midrule
\multirow{4}{*}{\rotatebox{90}{Simulation}} 
& CRV & 5.0 $\Leftrightarrow$ 76.0 & 34.1 $\Leftrightarrow$ 38.1 & 3.9 $\Leftrightarrow$ 88.3 & 20.7 $\Leftrightarrow$ 20.0 & 8.9 $\Leftrightarrow$ 80.7 & \textbf{71.9} \\
& CR  & 6.3 $\Leftrightarrow$ 71.3 & 34.4 $\Leftrightarrow$ 36.4 & 4.4 $\Leftrightarrow$ 81.7 & 21.6 $\Leftrightarrow$ 20.3 & 9.6 $\Leftrightarrow$ 73.7 & \textbf{64.1} \\
& RV  & 8.6 $\Leftrightarrow$ 31.9 & 15.1 $\Leftrightarrow$ 18.3 & 9.0 $\Leftrightarrow$ 54.3 & 6.3 $\Leftrightarrow$ 22.6 & 16.1 $\Leftrightarrow$ 55.4 & \textbf{39.3} \\
& R   & 8.4 $\Leftrightarrow$ 31.1 & 15.3 $\Leftrightarrow$ 18.6 & 9.9 $\Leftrightarrow$ 55.1 & 7.1 $\Leftrightarrow$ 23.4 & 16.1 $\Leftrightarrow$ 57.1 & \textbf{41.0} \\
\bottomrule
\end{tabular}
}
\caption{\textbf{Ablation models' win rate comparison} against $p_0$ across five evaluation dimensions and overall improvement (Imp.).}
\vspace{-.5cm}
\label{tab:ablation}
\end{table} 
We conduct two types of experiments to evaluate \ourmethod{} for planning: image-based procedural planning (\Sref{sec:results-procedural}) and embodied agent tasks (\Sref{section:results-embodied}). The image-based procedural planning experiments assess the effectiveness of \ourmethod{} in vision-language planning and provide insights into the types of self-reflection that are helpful for planning. We then evaluate directly on embodied agent tasks to demonstrate how \ourmethod{} results in direct improvements in vision-language procedural planning for embodied agents.

\paragraph{\ourmethod{} Implementation Details} We used a diverse range of base models to experiment with \ourmethod{}; Qwen-2.5-VL-Instruct (3B, 7B, 32B, 72B) \cite{bai2025qwen25vltechnicalreport} and Gemma 3 (4B, 12B, 27B) \cite{gemma}. Among open-sourced VLMs, these models have been shown to perform well on visual reasoning tasks \cite{cheng2024visionlanguagemodel}. 

Guided by a scaling experiment with number of revisions per round, we set the number of revisions to 2 for our main experiments. We set the number of maximum rounds to 5. 
For training, we set the temperature of the critique and refine stage to 0.5, while we use greedy decoding for the initial planning and validation stage.

We implement the \ourmethod{} as both an inference-time method (SelfReVision) and as supervised-finetuning (SelfReVision+SFT). For the SelfReVi+SFT method we curated a $n=160K$ subset of images from the \textsc{Places} Dataset \cite{zhou2017places}, which contains real-world scenes categorized by location type (e.g., airport lounge, kitchen, barn). We selected a diverse range of both indoor and outdoor scenes. Next, we used GPT-4o \cite{openai2024gpt4technicalreport} to generate a variety of plausible goals that a user might want to achieve in each given setting. Full experimental details are provided in \cref{appx:exp_detials}.
\vspace{-.2cm}

\subsection{Goal-Based Procedural Planning}
\label{sec:results-procedural}
\vspace{-.2cm}

\vspace{-.05cm}
\paragraph{Evaluation Dataset}
We evaluated \ourmethod{} on both real-world and simulation settings, as both settings frequently require procedural planning. For the real-world setting, we used a held-out test set of $n=100$ image and user-input pairs sampled from the \textsc{Places} Dataset \cite{zhou2017places}, and the corresponding user inputs were generated using GPT-4o \cite{openai2024gpt4technicalreport}.

For the \textsc{Simulation} setting, we used a modified version of the MFE-ETP dataset \cite{MFE-ETP}, which consists of $n=100$ image and user-prompt pairs drawn from the popular procedural simulation environments VirtualHome \cite{puig2018virtualhome} and BEHAVIOR-100 \cite{pmlr-v164-srivastava22a}. Since this dataset includes multi-image scenarios, we adjusted some user inputs to correspond to a single selected image when necessary. Example inputs and visualizations are shown in \cref{fig:data_example}, with additional details provided in \cref{appx:exp_details_procedural_planning}. 
\vspace{-.05cm}
\paragraph{Evaluation Metrics}
Prior work \cite{Brahman2023PlaSmaMS, osti_10366294} has evaluated procedural plans based on four dimensions: Coverage, Ordering, Completeness, and Overall Quality. We extend this framework by introducing a fifth criterion—Image Groundedness—to assess how well a plan aligns with the visual context. Specifically we define these criteria as: 
\begin{itemize}[leftmargin=*,noitemsep,topsep=0pt]
\item \textbf{Coverage}: How well the plan addresses the user's input.
\item \textbf{Ordering}: Whether the steps follow a logical and coherent sequence.
\item \textbf{Completeness}: Whether the plan is sufficiently detailed and informative.
\item \textbf{Image Groundedness}: Whether the plan is plausible given the visual scene.
\item \textbf{Overall Quality}: The overall effectiveness and appropriateness of the plan.
\end{itemize}

Given the strong performance of LLMs-as-judges \cite{10.5555/3666122.3668142}, we use GPT-4o \cite{openai2024gpt4technicalreport} as an automated evaluator via prompting. To validate this approach, we measured inter-rater reliability on a sample of $n=30$ and found an average agreement of $0.52$ between three GPT-4o judgements and three human annotators. This level of agreement is in line with the average agreement between humans. See \cref{appx:llm_as_judge} for full details. 

For our primary evaluation metric, we report the \textit{win rate}, which is the percentage of samples in which the revised plan (or model output) is preferred over that of the base model (i.e. $p_0$).

\vspace{-.05cm}
\paragraph{Baselines} 
To demonstrate the effectiveness of \ourmethod{}, we first compare the refined plans to the initial plans generated by the models using few-shot prompting. We also evaluate responses from other baselines such as GPT-4o (representing a powerful large model) \cite{openai2024gpt4technicalreport}, PaliGemma (a domain-specific model trained for planning) \cite{beyer2024paligemmaversatile3bvlm}, and best-of-N (an inference-time algorithm that generates multiple outputs and selects the best one). The prompts and examples provided for GPT-4o and PaliGemma match those given to the base models.
For the best-of-N baseline, we use $N\!=\!5$: we sample five different plans with a temperature of 0.5, followed by a final inference step to select the best plan among them. This setup approximately matches the number of additional inferences made by both \ourmethod{} and the baseline.

\vspace{-.05cm}
\paragraph{Results: \ourmethod{} yields large and consistent improvements over baselines.} 
\cref{tab:full-results} shows that across all model sizes and both datasets, \ourmethod{} consistently outperforms the initial plans $p_0$ by wide margins. Specifically, there is an average win rate $68\%$ on \textsc{Places} and $72\%$ on \textsc{Simulation}, with the most dramatic gains in completeness and coverage—often surpassing $80\%$ win rates against the base plans. These results demonstrate that iterative self-improvement through \ourmethod{} is highly effective in enhancing the structure, richness, and plausibility of plans, regardless of model size. Notably, larger models tend to benefit even more from \ourmethod{}, both in absolute win rates and in the consistency of gains across metrics. For example, models over 12B have on average $74\%$ gain overall using \ourmethod{} compared to $68\%$ for models 12B and under.

Compared to alternative methods such as \textsc{Best-of-N} sampling and PaliGemma, \ourmethod{} shows clear superiority. While Best-of-N offers modest improvements for small models ($8\% - 38\%$), \ourmethod{} provides substantially higher gains ($60\%$ across most settings). Somewhat unexpectedly, PaliGemma—a strong pretrained VLM—consistently underperforms, losing over $90\%$ of matchups across both datasets. Despite being trained on image distributions similar to those in \textsc{Places}, it appears to lack the procedural reasoning abilities required for grounded multi-step planning, suggesting its limitations in this domain.

Lastly, we also assess the impact of SFT on \ourmethod{} outputs. While SelfReVision+SFT achieves moderate gains in some settings (e.g., $57\%$/$54\%$ for Qwen-7B in \textsc{Places} and \textsc{Simulation}), it sometimes underperforms compared to using the raw output, and in several cases yields no improvement. This suggests that while fine-tuning can help stabilize refinement behavior, it may also dilute some of the benefits of the iterative reasoning process when not tuned carefully. 

\begin{figure}[t]
    \centering
    \includegraphics[width=\linewidth]{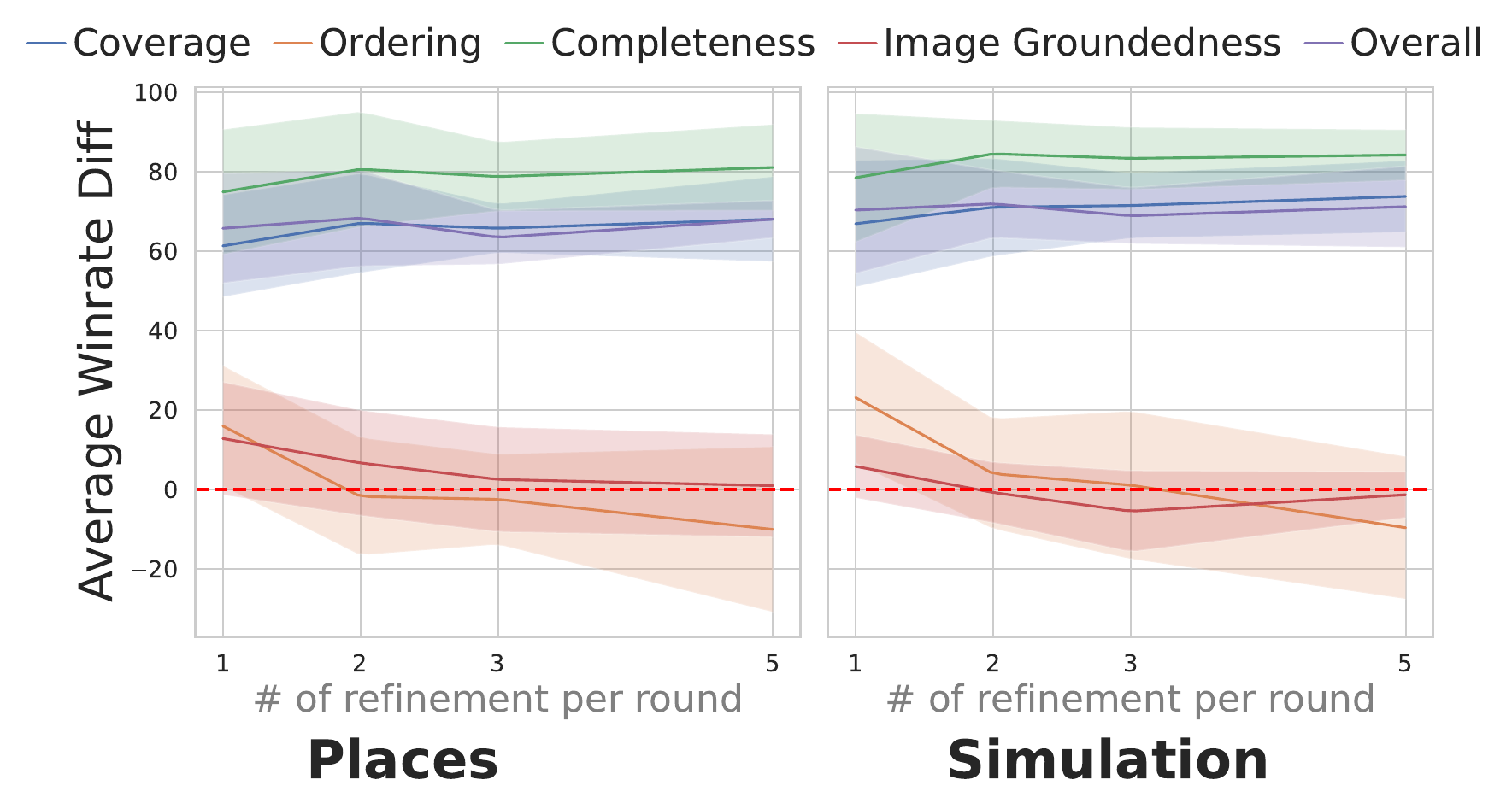}
    \caption{\textbf{Average winrate difference} (winrate of SelfReVision -  $p_0$) over number of refinement per round.}
    \label{fig:scaling}
    \vspace{-17pt}
\end{figure}

\hspace{-2em}
\begin{figure*}[ht]
    \centering
\
    \begin{subfigure}[t]{0.53\textwidth}
        \centering
        \renewcommand{\arraystretch}{1.4}
        \setlength{\tabcolsep}{6pt}
        \begin{tabular}{@{} >{\raggedright\arraybackslash}p{3cm} c c c @{}}
        \toprule
        \textbf{Goal} & \textbf{Initial State} & \textbf{$P_0$} & \textbf{SelfReVision} \\
        \midrule
        Create a smiley face. & 
        \includegraphics[width=2.2cm]{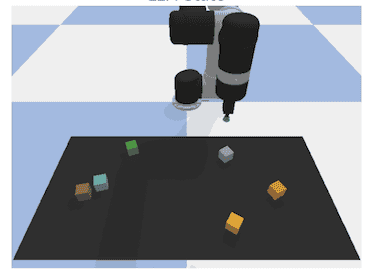} & 
        \includegraphics[width=2.2cm]{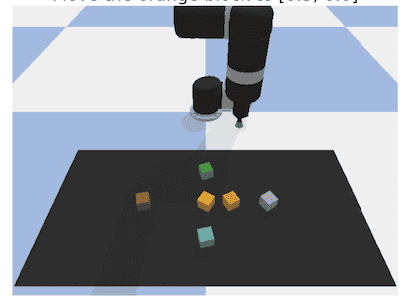} & 
        \includegraphics[width=2.2cm]{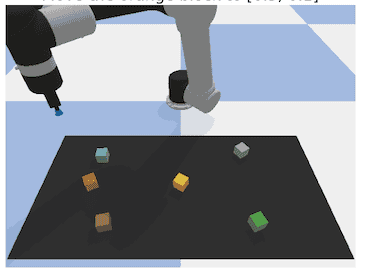} \\

        Form a rainbow. & 
        \includegraphics[width=2.2cm]{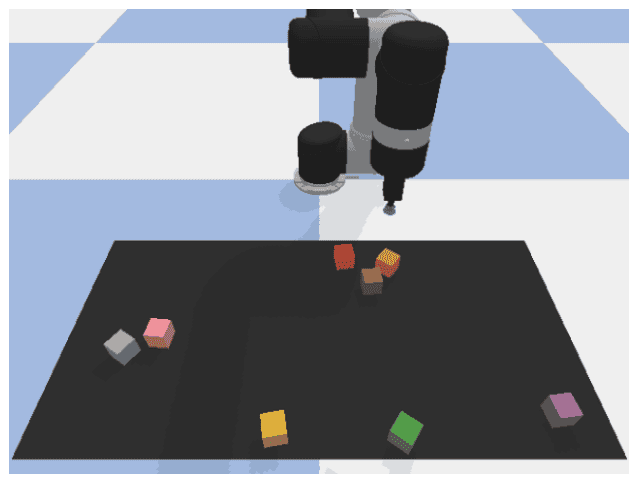} & 
        \includegraphics[width=2.2cm]{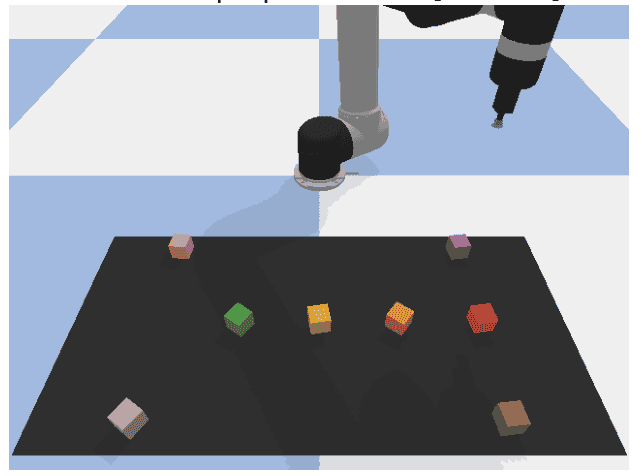} & 
        \includegraphics[width=2.2cm]{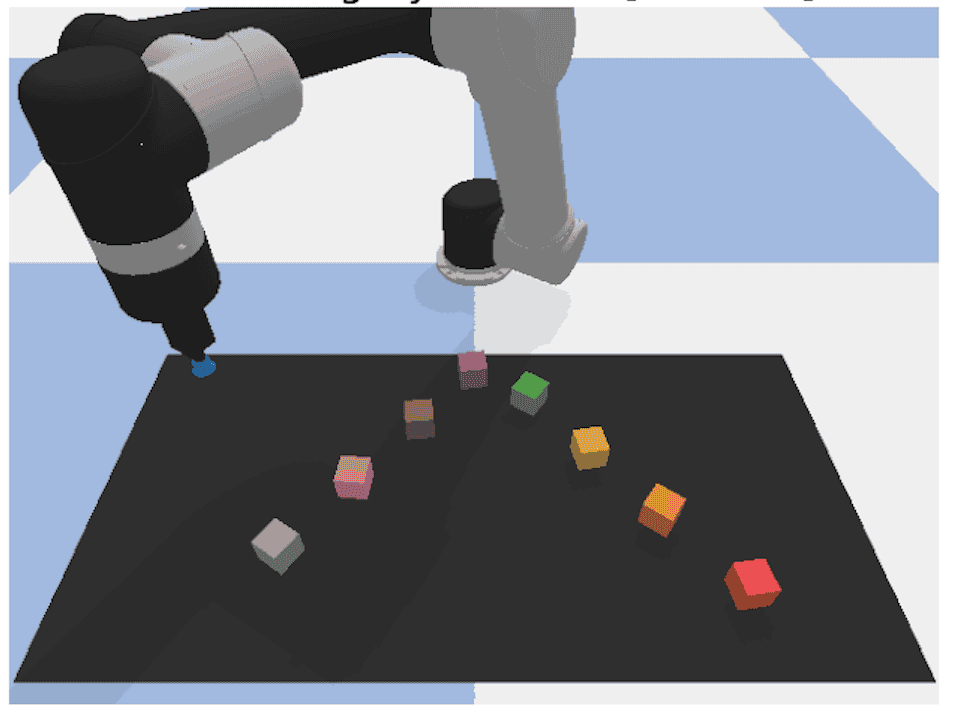} \\

        Form an uppercase O. & 
        \includegraphics[width=2.2cm]{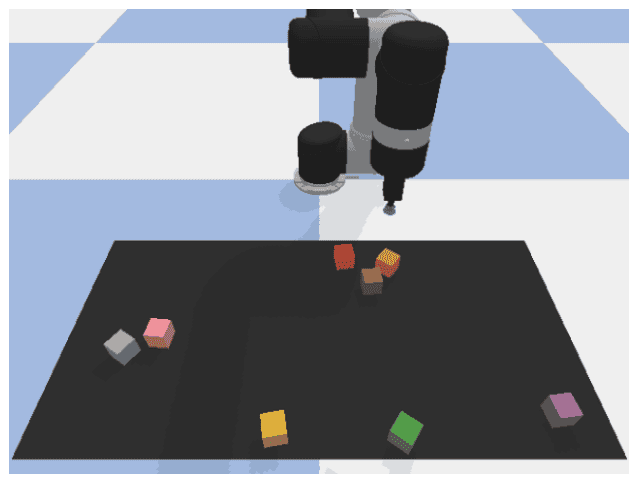} & 
        \includegraphics[width=2.2cm]{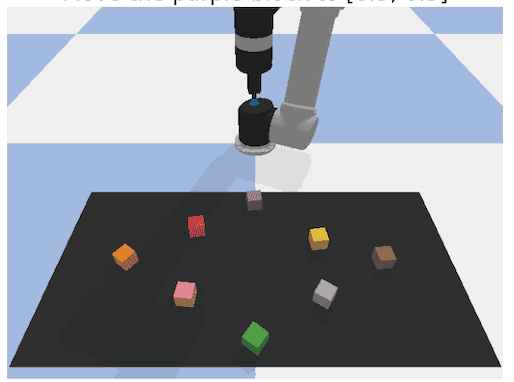} & 
        \includegraphics[width=2.2cm]{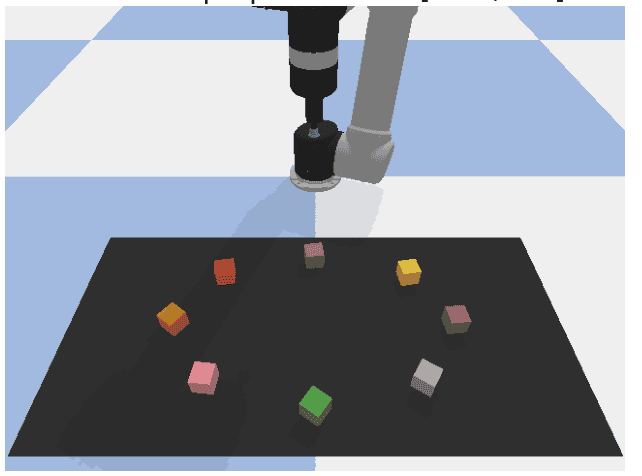} \\
        \bottomrule
        \end{tabular}
        \caption{Block-building goals, initial state, $P_0$, and SelfReVision outputs.}
        \label{fig:block_example}
    \end{subfigure}
    \hfill
    \begin{subfigure}[t]{0.3\textwidth}
        \centering
        \vspace*{-10.2em} 
        \includegraphics[width=.65\linewidth]{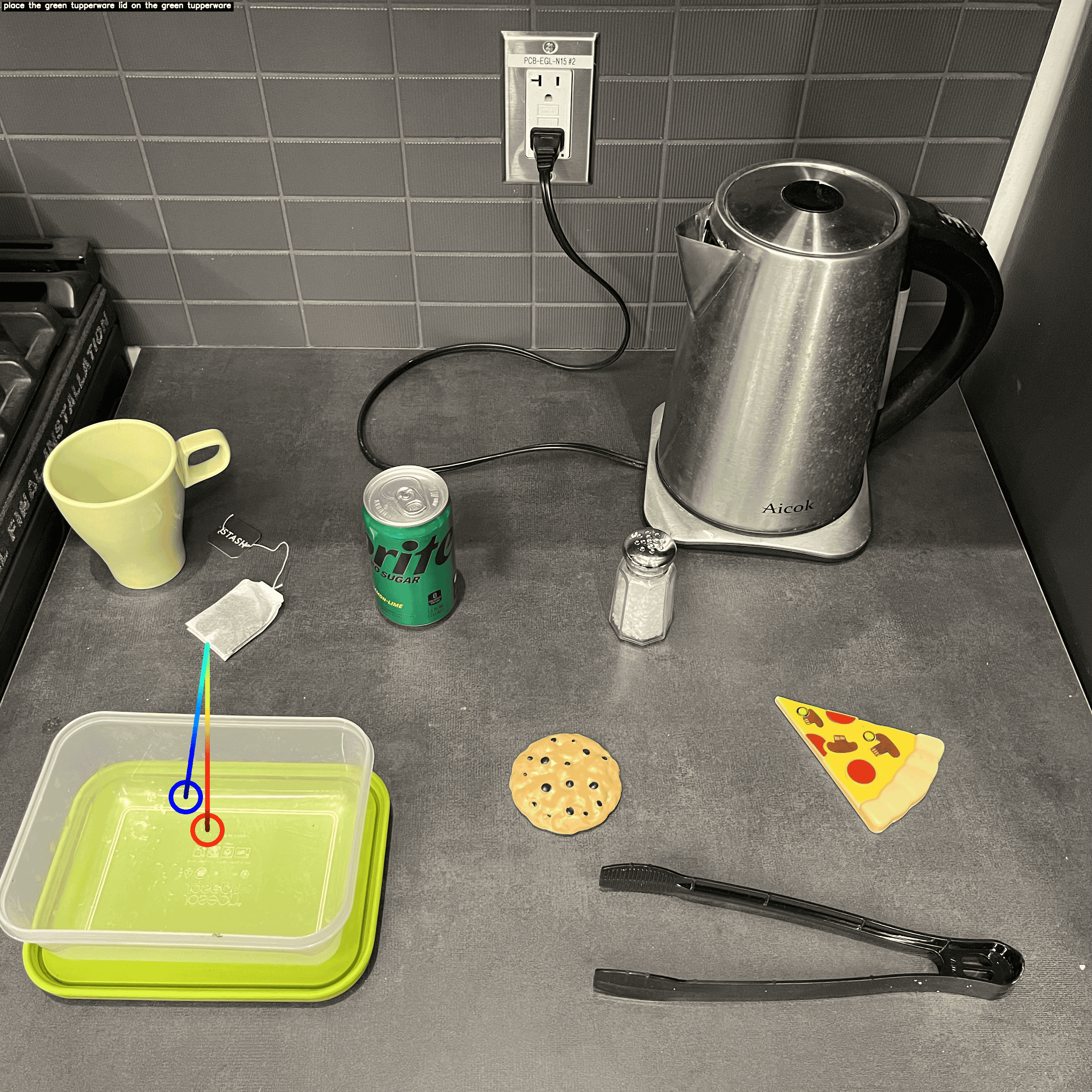}
        \caption*{\textbf{Addition:} \textit{"place the green tupperware lid on the green tupperware"}}
        \includegraphics[width=.65\linewidth]{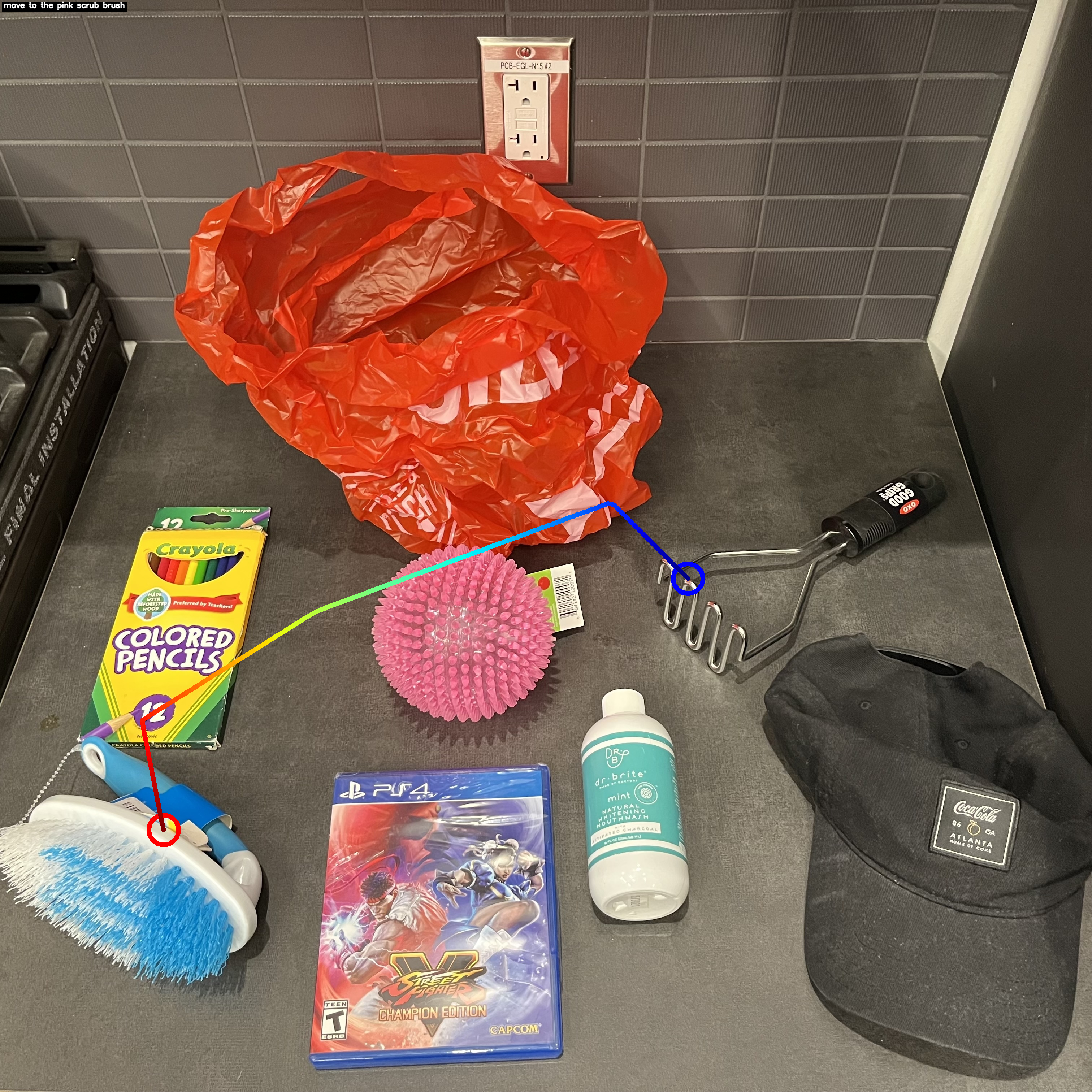}
        \caption*{\textbf{Removal:} \textit{"pick up and place the pink scrub brush into the red bag"}}
        \caption{Object manipulation with SelfReVision in hierarchical planning.}
        \label{fig:hamster_changes}
    \end{subfigure}

    \caption{\textbf{Examples from two embodied agent tasks}: (a) block-building goals, initial setting, and then finalized setting after running $P_0$, and SelfReVision plans. The first two rows show examples from Gemma 12B and the last row is from Gemma 27B.; (b) examples of correct addition and removal of SelfReVision plan in hierarchal planning.}
    \vspace{-.25cm}
    \label{fig:final_combined}
\end{figure*}
 \vspace{-.05cm}
\paragraph{Results: \ourmethod{} produce better plans than GPT\-4o.} To assess how \ourmethod{} stacks up against significantly larger models, we compare the win rate of plans it generates with those produced by GPT\-4o, as shown in \cref{tab:gpt4o}. Our results reveal that for models with 12B parameters or more, \ourmethod{} achieves a win rate at least $25\%$ higher than GPT-4o. This highlights the effectiveness of self-critical, self-revision strategies in enabling even smaller models to outperform much larger ones.

\vspace{-.05cm}
\paragraph{Results: Tradeoffs in Refinement Scaling} 
We examined how \ourmethod{}'s performance changes with more refinement cycles in its self-refinement loop. As shown in \cref{fig:scaling}, the average Overall win rate rises from $75\%$ to $81\%$ on \textsc{Places} and from $78\%$ to $81\%$ on \textsc{Simulation} as the number of rounds increases from 1 to 5. However, the gains vary by metric: Coverage and Completeness steadily improve (e.g., +11 and +10 on \textsc{Places}), suggesting that additional rounds help produce more thorough plans. In contrast, Ordering and Image-Groundedness decline slightly (–5 and –3), indicating that later rounds may introduce speculative or less visually anchored content. Early refinements tend to add useful specifics (e.g., ``$80\%$ fill''), while later ones often bring more tentative phrasing (e.g., ``if there is water in the cup''), reflecting a trade-off between elaboration and precision. Notably, most of the improvement occurs within the first 2–3 rounds, showing that a few iterations are often enough to achieve strong results without sacrificing clarity.

\vspace{-.05cm}
\paragraph{Results: Ablation of Pipeline Steps}
To evaluate the contribution of each component in \ourmethod{} self-refinement loop, we conducted a series of ablation experiments by selectively removing individual stages. Table~\ref{tab:ablation} presents the ablation results on both the \textsc{Places} and \textsc{Simulation} datasets, averaged across the seven VLMs. We compare four configurations: the full CRV (Criticize-Revise-Verify) pipeline, CR (Criticize-Revise), RV (Revise-Verify), and R (Revise-only).
The details on ablation model variants can be found in \Aref{appx:exp_details_procedural_planning}.

The full CRV pipeline yields the strongest performance, with average win-rate improvements of 68.3\% on the \textsc{Places} dataset and 71.9\% on the \textsc{Simulation} dataset. This result confirms that integrating all three stages produces the most robust improvements in procedural plan quality. Notably, compared to CR, we observe significantly larger performance drops with RV and R. These variants especially show reduced improvements in Coverage and Completeness, indicating the essential role of the Criticize step in generating more comprehensive plans that better address user requests. 

While the CR variant demonstrates the best performance among the ablated configurations, it still exhibits notable performance drops (-9.3\% on \textsc{Places} and -7.8\% on \textsc{Simulation}) relative to the full CRV. In some cases, the refined plans were even worse than the initial plans in terms of Ordering and Image Groundedness. These results suggest that the Verify step plays a critical role in filtering out suboptimal revisions--particularly those that disrupt the correct order or misalign with visual context. Together, these findings underscore that each stage in \ourmethod{} contributes distinct and complementary benefits to plan refinement.

\begin{figure*}[t]
    \centering
    \includegraphics[width=0.9\linewidth]{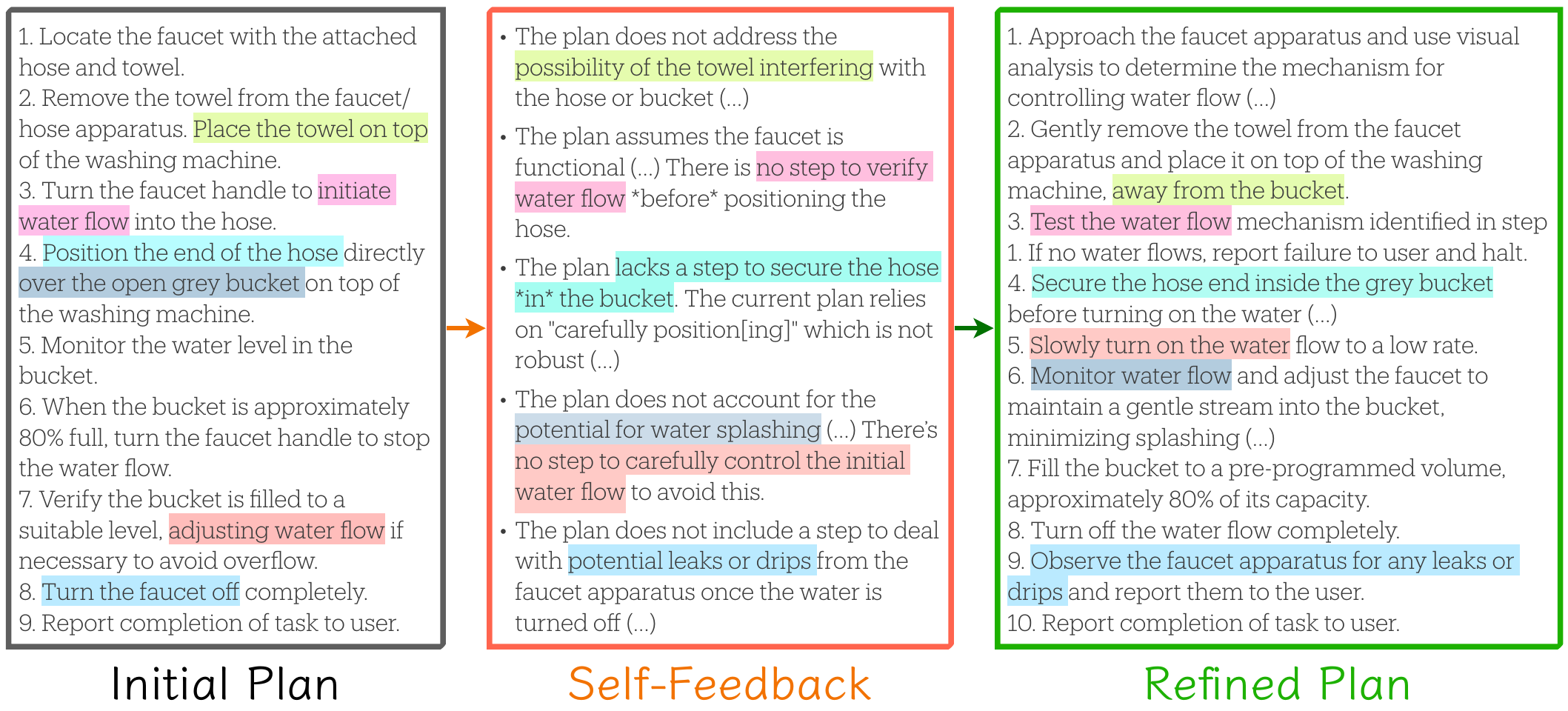}
    \caption{Initial plan, feedback, and refined plan generated by Gemma-27B for the example in Figure 1.}
    \label{fig:example}
    \vspace{-.5cm}
\end{figure*}
\vspace{-.05cm}
\paragraph{Qualitative Analysis}
\cref{fig:example} provides an example of an initial plan generated by Gemma-27B, along with the self-feedback and refined plan produced by \ourmethod{}. Although the initial plan seems sufficiently clear at first glance, the self-critique step identifies critical shortcomings such as positioning the hose after turning on the water and potential interference by placing the towel too close to the bucket. The refined plan explicitly addresses these issues (e.g., ``place the towel away from the bucket,'' ``secure the hose end inside the bucket before turning on the water''). Additionally, the refined version includes explicit instructions regarding robot-specific considerations-- monitoring for leaks or splashes-- details intuitive to humans but essential for robotic execution. This iterative refinement thus results in a more robust and executable plan.

\vspace{-.2cm}
\subsection{Application to Embodied Agents}
\vspace{-.2cm}

\label{section:results-embodied}
To study the ability of \ourmethod{} to improve planning in embodied settings, we construct two challenging scenarios: (1) a simulated pick-and-place environment~\cite{zeng2020transporter} controlled by code-as-policies~\cite{liang2023code} and (2) a real-world planning environment based on path-prediction inspired by HAMSTER~\cite{li2024hamster}. We limit our evaluation to the models that were best at baseline procedural planning, Gemma 12B and 27B \cite{gemma}.

\vspace{-.05cm}
\paragraph{Evaluation Dataset}
For the simulated pick-and-place enviornments, we first curated 14 semantically unique manipulation goals (e.g., ``Form a shape of an uppercase X with the blocks'', ``Create a smiley face'') and paired them with 8 different initial block configurations involving 6 or 8 blocks from \citet{zeng2020transporter}. This yielded a total of $n=112$ samples. For the real-world setting, we created 10 real scenarios across three environments -- kitchen, workshop, and office -- each involving a high-level task (e.g., “Pack items for a children’s lunch”). 

\vspace{-.05cm}
\paragraph{Evaluation Metrics}
For the simulated pick-and-place enviornment, we ran each plan using a code-as-policies simulator \cite{liang2023code} which generated a static image for each step. Then, a human rater evaluated the final configuration, judging whether the plan achieved the stated goal.
For the real-world settings, we used \citet{li2024hamster} to generate a trace path for each step in each generated plan. Then, a human raters assessed whether each individual step was completed successfully by the generated trace.

\begin{table}[]
\resizebox{\columnwidth}{!}{
\begin{tabular}{lrr|rr|}
         & \multicolumn{2}{c}{\textbf{Gemma-12b}}                                        & \multicolumn{2}{c}{\textbf{ Gemma-27b}}                                        \\
         \cline{2-5}
      \multicolumn{1}{l|}{}   & \multicolumn{1}{l}{\text{$p_0$}} & \multicolumn{1}{l|}{\text{SelfReVis.}} & \multicolumn{1}{l}{\text{$p_0$}} & \multicolumn{1}{l|}{\text{SelfReVis.}} \\
         \hline
\multicolumn{1}{|l|}{6 Blocks} & 0.16                            & \textbf{0.45}                         & 0.36                            & \textbf{0.59}                         \\
\hline
\multicolumn{1}{|l|}{8 Blocks} & 0.14                            & \textbf{0.39}                         & 0.29                            & \textbf{0.39}      \\
\hline
\end{tabular}
}
    \caption{Results of the simulated block manipulation tasks, showing the average success rate for both the baseline $p_0$ and SelfReVision plans on settings with 6 or 8 blocks over Gemma 12B and 27B.}
    \label{tab:block_results}
    \vspace{-.5cm}
\end{table} \vspace{-.05cm}
\paragraph{Results: \ourmethod{} improves downstream performance on block manipulation task and real-world planning scenarios} As shown in \cref{tab:block_results}, the plans enhanced by \ourmethod{} outperformed the base model plans by $26\%$ (12B) and $17\%$ (27B), respectively. Qualitatively, the improvements were especially notable in more complex tasks like "Create a smiley face" or "Form a rainbow". For the smaller 12B model, \ourmethod{} often transformed failed attempts into successful plans (see \cref{fig:block_example}). In contrast, for the larger 27B model, the improvements were more subtle—enhancing already successful outputs, such as making the structure more rounded in the final example of \cref{fig:block_example}. These results indicate that the critical revision process introduced by \ourmethod{} can produce higher-quality plans that more reliably complete manipulation tasks.

For the hierarchical task, we found that the \ourmethod{} plans resulted in $70\%$ successful traces creation by the HAMSTER action model compared to only $61\%$ of the base model plans. These improvements stemmed from both meaningful additions and removals within the plans, resulting in more accurate downstream traces. \Cref{fig:hamster_changes} presents two illustrative examples of such revisions and their downstream impact. In the top image, where the goal was to pack a kid's lunch, \ourmethod{} correctly added a final missing step to place the lid on the Tupperware. In contrast, the bottom image shows an error in the base plan for the goal ``pack toys for a kid,'' where the model mistakenly included an action involving a blue scrub brush, misidentifying it as a toy. \ourmethod{} successfully removed this unnecessary step. These examples highlight how \ourmethod{} enhances plan precision by correcting both omissions and errors, leading to more reliable task execution.

\paragraph{Conclusion}
We showed that \ourmethod{}, a self-improvement framework for vision-language procedural planning, can significantly boost the performance of small models through iterative self-critiquing and refinement. %
\section*{Limitations}
While our method demonstrates promising results for low-level procedural planning in small-scale VLMs, it is not without limitations.

A primary limitation of \ourmethod{} is its increased inference cost. Unlike the SFT approaches that generate a complete plan in a single forward pass, \ourmethod{} requires iterative refinement across multiple calls—averaging around 8 inference steps per example. This iterative process enables more accurate and grounded reasoning, but may pose challenges for latency-sensitive or real-time applications.

Second, our self-improvement strategy assumes that the model can recognize and correct its own planning errors during training. However, if the model’s internal reward signal or critique mechanism is flawed, this could reinforce incorrect behaviors or lead to overfitting on superficial plan heuristics. Although we do see improvement in all models tested, a weaker model might not benefit from the same method.

Lastly, currently we only experiment with added visual inputs and do not incorporate other potentially useful modalities such as robot proprioception, or tactile feedback. This unimodal design limits the method’s ability to adapt to multimodal real-world scenarios where contextual or embodied cues are critical for accurate planning. It would be interesting for future work to attempt to incorporate more versatile type of information in the self-critiquing loop. 
 \paragraph{Acknowledgement} This work was upported by funding from the Army research lab. 
\newpage
\bibliography{custom}
\newpage
\appendix
\section{Experimental Details}\label{appx:exp_detials}
In this section, we provide full details of the experimentation used in this paper. We start with implementation of our method \cref{appx:our_method_implement}, and then discuss the experimental setup of both the procedural planning \cref{appx:exp_details_procedural_planning} and embodied agents \cref{appx:exp_details_embodied_agents}.

\subsection{Method Implementation} \label{appx:our_method_implement}
\paragraph{Training Data}\label{appx:exp_details_training_data}
We used a subset of images from the Places365 Dataset \cite{zhou2017places}, which contains real-world scenes categorized by location type (e.g., airport lounge, kitchen, barn). This dataset originally contains 2.5 million images which are categorized into 205 types of scenes (e.g. barn, living room, beauty salon). Some of these categories were not conducive to our experiments, specifically ones that might not allow for many tasks to be done (e.g. barndoor, batters box, ice shelf). To determine which categories to use, we had two researchers independently rate all 205 categories based on perceived eligibility to the task of procedural planning on a 4-point likert scale (1 = best category, 4 = worst category). We then included all categories which had an average score of 1.5. This resulted in the following diverse 55 categories:
\begin{itemize}
    \item \textbf{Places365 Categories}: airplane cabin, airport terminal, apartment building outdoor, aquatic theater, arcade, archaeological excavation, archive, army base, art gallery, art studio, atrium public, banquet hall, bar, barn, basement, bathroom, bazaar indoor, beach house, biology laboratory, bookstore, chemistry lab, childs room, classroom, clothing store, coffee shop, dinette home, dorm room, florist shop indoor, florist shop outdoor, gallery, game room, gymnasium indoor, hardware store, home office, home theater, hospital, hospital room, hotel room, kindergarten classroom, kitchen, kitchenette, laundromat, living room, lobby, nursery, office, pharmacy, playroom, pub indoor, reception, recreation room, repair shop, restaurant kitchen, storage room, utility room. 
\end{itemize}
We aimed to based our dataset on a diverse range of real-world images, including both indoor and outdoor scenes. 

Then, within each category there is a wide range of types of images. Since this dataset uses images from a wide range of online sourced, not all the images are of the same quality. For our task, we wanted to have scenes which were clear, easy to see, and not too focused on one object or too broad to not be able to have tangible tasks. Therefore, we choose to filter the images based on the following criteria:
\begin{itemize}
    \item \textbf{Too Blurry}: slight blurriness is acceptable if objects remain identifiable, but excessively blurry images should be excluded.
    \item \textbf{Too Dark}: some darkness is acceptable as long as objects can still be discerned. However, images that are too dark to identify objects should be filtered out.
    \item \textbf{Too Zoomed-In/Too Zoomed-Out}: images that are overly focused on a single detail (e.g., close-ups of flowers or a single individual) and lack broader environmental context should be excluded./images taken from too far away, like more than 100 feet away, or those that primarily capture abstract landscapes, making it difficult to infer meaningful tasks specific to the environment, should be filtered out
\end{itemize}
We did this filtering automatically using GPT-4o \cite{openai2024gpt4technicalreport} by prompting. The exact prompt can be see in \cref{appx:filter_prompt}. In total we randomly selected 51997 (1000 images per category) images, resulting in 35619 final images after filtering.

Next, we took each of these filtered images and again prompted GPT-4o to generate a plausible user-input (see \cref{appx:generate_user_input_prompt}). This resulted in a final dataset of $n=107013$ image/user-input pairs for training.

\begin{prompt}\label{appx:filter_prompt}
You are evaluating an image to decide whether it should be filtered out for data generation purposes.
An ideal image should provide clear environmental context for robots, as these images will be used to generate a list of tasks that robots can perform based on the given situation.
Specifically, images should be filtered out if they meet any of the following criteria:
1) too blurry (slight blurriness is acceptable if objects remain identifiable, but excessively blurry images should be excluded.),
2) too dark (some darkness is acceptable as long as objects can still be discerned. However, images that are too dark to identify objects should be filtered out.),
3) too zoomed-in (images that are overly focused on a single detail (e.g., close-ups of flowers or a single individual) and lack broader environmental context should be excluded.),
4) too far-out (images taken from too far away, like more
 than 100 feet away, or those that primarily capture abstract landscapes, making it difficult to infer meaningful tasks specific for the environment, should be filtered out).

Please provide feedback for each criterion and the overall decision in JSON format as shown in the example below:
`{"blurry":"blurry/ok","darkness":"too dark/quite dark/slightly dark/ok", "zoomed-in": "too zoomed-in/somewhat zoomed-in/ok", "far-out":"too far-out/somewhat far-out/ok", "decision":"keep/filter"`}
\end{prompt}

\begin{prompt}\label{appx:generate_user_input_prompt}
Given an image generate 3 plausible user inputs from someone in the image directed at a robot, which would then cause the robot to do a task. The user inputs can be statements or questions. 

Also, for each input, generate a list of high-level steps for the robot to finish the task. Make sure the high-level steps are specific to the setting in the image.

Lastly, for each input, generate a short response by the robot that indicates what it plans to do.

Do not mention the image or picture. The user inputs should be very different from each other and specific to the scene. Separate the high-level steps using "|". Respond strictly in JSON format with 9 keys: ‘User\_Input1’, ‘Steps1’, ‘Robot\_Response1’,…, ‘User\_Input3’, ‘Steps3’, ‘Robot\_Response3’. Do not use any markdown formatting or code block symbols (such as triple backticks).

** Multiple like-version of this prompt was used, see Github code for full list**
\end{prompt}

\paragraph{Self-Distillation/Improvement} In order to generate the high-level plans (the labels of our training data) we used the base model itself through prompting in a process called self-distillation. First, we use a general prompt to get an initial plan $p_0$ (see \cref{appx:initial_prompt}. 

However, given the weak nature of the base model, this prompt is not going to be well grounded to the given scene. Therefore, we use a series of self-critique, self-revise, and self-evalute prompts to generate a better final plan. First, we self-critique the initial plan using an open-ended prompt Crit$(p_0)$, see \cref{appx:critique_prompt}. Then, we used the output from this prompting along with the original plan $p_0$ to revise the original plan Rev$(p_0,$Crit$(p_0))= p_1$, see \cref{appx:revise_prompt}. Lastly, we prompted the base model to verify if the revised plan is better than the original plan using \cref{appx:verify_prompt} $\text{Ver}(p_0, p_1)$.

\begin{prompt}\label{appx:initial_prompt}
You are writing instructions for a robot in the image. Make a detailed plan which responds to the users input. You can only use the items you see in the given image and must make your plan specific to this setting.

You should respond with only the numbered plan which starts with ``<plan>'' and ends with ``</plan>''.
No other text should be outputted. Do not use any markdown formatting, code block symbols (such as triple backticks), headings, summaries, or nested bullet points

User Input: ``\{user\_input\}''
\end{prompt}

\begin{prompt}\label{appx:critique_prompt}
You are reviewing a high-level plan for a robot based on a user request and an image of the environment.

Your goal is to identify critical flaws, gaps, or missed opportunities that would significantly improve the plan’s feasibility, clarity, or alignment with the depicted environment. Focus on major missing steps, unrealistic assumptions, or vague actions that reduce the quality of the plan. Avoid nitpicking or commenting on minor stylistic issues.

Ground your feedback in the visual context and user intent. Prioritize issues that would materially impact the robot’s ability to execute the task successfully.

Output a clean, single-level numbered list of feedback enclosed between <critic> and </critic>. Each item should describe one clear issue or suggestion for meaningful improvement.

Do not suggest rewordings or edits—focus only on diagnosing problems.

User Input: ``\{user\_input\}''

Current Plan: ``\{current\_plan\}''
\end{prompt}

\begin{prompt}\label{appx:revise_prompt}
You are revising a high-level robot plan based on critical feedback, the user’s request, and an image of the environment.

Use the feedback to identify key flaws and address them with substantive improvements. Focus on clarity, feasibility, and grounding the plan in the actual visual context. Prioritize corrections that enable the robot to effectively and realistically complete the task.

Make **meaningful changes**, not surface-level edits. Omit redundant or overly detailed instructions that don't improve execution. Avoid speculative details unless they're clearly justified by the visual context.

Output a clean, single-level numbered list of steps enclosed between <plan> and </plan>. Do not include titles, nested lists, extra commentary, or any formatting besides the numbering.

User Input: ``\{user\_input\}''

Current Plan: ``\{current\_plan\}''

Feedback: ``\{criticism\}''
\end{prompt}

\begin{prompt}\label{appx:verify_prompt}
You are evaluating two sets of instructions for a robot in the image.
You will be given a user input and two high-level plans.
Compare the two plans and respond with "yes" if Plan 2 better fulfills the user request than Plan 1; otherwise, respond with "no".
Good plans generally use only items visible in the image and are specific to the setting shown.
A better plan more effectively uses only items visible in the image and is more specific to the setting shown.
It also demonstrates stronger coverage, more logical order, greater completeness, and better grounding in the image.
Do not use any markdown formatting or code block symbols (such as triple backticks).''
                  
User Input: ''\{user\_input\}''

Plan 1: ''\{initial\_plan\}''

Plan 2: ''\{revised\_plan\}''
\end{prompt}

\paragraph{Training}
\label{appx:training_details} We used a diverse range of base models to experiment with \ourmethod; Qwen-2.5-VL-Instruct (3B, 7B, 32B, 72B) \cite{bai2025qwen25vltechnicalreport} and Gemma 3 (4B, 12B, 27B) \cite{gemma}. 
We performed supervised fine-tuning of the base models using plans generated with \ourmethod. During training, all models were cast to the torch.bfloat16 data type and trained for 4 epochs. The best model was selected based on cross-entropy loss on a development set consisting of 100 randomly held-out examples from the training data. Final evaluation results (win rates) were computed on a separate set of 100 held-out samples. We experimented with three learning rates (1e-5, 3e-5, and 5e-5) for each model and report results for the best-performing one. Weight decay was fixed at 0.01, and the maximum number of tokens was set to 500 for all models.

\subsection{Goal-Based Procedural Planning Details}\label{appx:exp_details_procedural_planning}
In this section we outline the experimental details for the goal-based procedural planning experiments. 

\paragraph{Evaluation Dataset}  We evaluated our method on both real-world setting and simulation setting datasets. For the real-world setting, we used a randomly selected held-out test set of $n=100$ image and user-input pairs from our training data. These images were sampled from the Places365 Dataset \cite{zhou2017places}, and the corresponding user inputs were generated using GPT-4o \cite{openai2024gpt4technicalreport}. See \cref{appx:exp_details_training_data} for full details.

For the simulation setting, we used a modified version of the MFE-ETP benchmark dataset \cite{MFE-ETP}, which consists of $n=100$ image and user-prompt pairs drawn from the popular procedural simulation environments VirtualHome \cite{puig2018virtualhome} and BEHAVIOR-100 \cite{pmlr-v164-srivastava22a}. This dataset was created as a challenging benchmark for embodied reasoning and procedural planning. However, for some of the original MFE-ETP samples, there are multiple images of the initial conditions which might be needed to create a plan for the given task. Since, we want to focus on only one image for a user-input, we hand-selected teh best image for the given task. If no image captured enough information to complete the task, we randomly selected an image and wrote a new task. The full list of the $n=100$ choosen images and tasks can be found on our github. 

\paragraph{Baselines}
To demonstrate the effectiveness of \ourmethod{}, we first compare the refined plans to the initial plans generated by the models using few-shot prompting. We also evaluate responses from other baselines such as GPT-4o (representing a powerful large model) \cite{openai2024gpt4technicalreport}, PaliGemma (a domain-specific model trained for planning) \cite{beyer2024paligemmaversatile3bvlm}, and best-of-N (an inference-time algorithm that generates multiple outputs and selects the best one). The prompts and examples provided to GPT-4o and PaliGemma match those given to the base models.
For the best-of-N baseline, we use $N\!=\!5$: we sample five different plans with a temperature of 0.5, followed by a final inference step to select the best plan among them. This setup approximately matches the number of additional inferences made by both \ourmethod{} and the baseline.

\cref{appx:baseline_fewshot_prompt} shows the exact prompt used to do few-shot generation with baselines. 

\begin{prompt}\label{appx:baseline_fewshot_prompt}
You are writing instructions for a robot in the image. Make a detailed plan which responds to the users input.
You can only use the items you see in the given image and must make your plan specific to this setting.
You should respond with only the numbered plan and no other text should be outputted.
Do not use any markdown formatting or code block symbols (such as triple backticks).

Example 1
User Input: Hmm, I don't think the time on that clock is correct.
Plan: 1. Navigate to the Clock
2. Grab the Clock
3. Adjust the Time to 12:15
4. Return the Clock

Example 2
User Input: Can you make my drink colder?
Plan: 1. Navigate to the Fridge
2. Open the Freezer Door
3. Locate the Ice Tray
4. Collect the Ice
5. Close the Freezer Door
6. Navigate back to the Person
7. Put the Ice in the Drink

Example 3
User Input: Can you hang this picture for me?
Plan: 1. Pick up the Hammer and Nail
2. Insert Nail into the Wall with Hammer
3. Put Down the Tools
4. Pick up Picture
5. Hang the Picture

User Input: {user\_input}
Plan:
\end{prompt}

\paragraph{Ablation Study Details}
To evaluate the contribution of each component in \ourmethod{} self-refinement loop, we conducted a series of ablation experiments by selectively removing individual stages. Table~\ref{tab:ablation} presents the ablation results on both the \textsc{Places} and \textsc{Simulation} datasets, averaged across the seven VLMs. We compare four configurations: the full CRV (Criticize-Revise-Verify) pipeline, CR (Criticize-Revise), RV (Revise-Verify), and R (Revise-only).

\cref{appx:ablation_prompt_revision_without_feedback} shows the revision prompt for variants that do not go through the self-criticism process (RV and R). 

\begin{prompt}\label{appx:ablation_prompt_revision_without_feedback}
You are revising a high-level plan for a robot. You will be given a user's input and the current plan. Your task is to revise and improve the plan. 

When revising:
1. Make sure to use only objects visible in the image
2. Provide a step-by-step plan specific to the setting
3. Address all aspects of the user input
4. Ensure logical ordering of actions
5. Add spatial details where needed
6. Ensure all actions are feasible in the environment shown

Respond only with the revised, numbered steps which starts with "<plan>" and ends with "</plan>". Do not include any additional text. Do not use markdown formatting or code block symbols (such as triple backticks).

User Input: {user\_input}
Current Plan: {current\_plan}
\end{prompt}

\paragraph{Evaluation Methodology and Other Details}
In line with prior work \cite{Brahman2023PlaSmaMS, osti_10366294}, we evaluate procedural plans using the following five criteria:

\begin{itemize} 
\item \textbf{Coverage} — How well the plan addresses the user's input. 
\item \textbf{Ordering} — Whether the plan follows a coherent and logical sequence. 
\item \textbf{Completeness} — Whether the plan is sufficiently detailed and informative. 
\item \textbf{Image Groundedness} — Whether the plan is plausible given the specific visual scene. 
\item \textbf{Overall Quality} — The overall effectiveness and appropriateness of the plan. 
\end{itemize}

We include the \textit{Image Groundedness} criterion to reflect the visual nature of our model: unlike prior work focused solely on language models (LLMs), our goal is to develop a vision-language model (VLM) that generates plans tailored to specific images.

Given the strong performance of LLMs-as-judges \cite{10.5555/3666122.3668142}, we use GPT-4o \cite{openai2024gpt4technicalreport} as an automated evaluator via prompting. See \cref{appx:llm_as_judge} for full details on validation of this method. The prompt we used to evaluate can be seen in \cref{appx:eval_prompt}.

\begin{prompt}\label{appx:eval_prompt}
You will be given an image of a setting, a user input and a corresponding plan with high-level steps that can be used by a robot to respond to the user input in that setting. Only output a valid json (python dictionary) and keep any explanation brief < 10 words.
    Your task is to evaluate the plan based on the following five criteria:\\

     Coverage (Does the plan fully address the user input?)**\\
    - **5 (Definitely): The plan thoroughly addresses all aspects of the user input without omissions.\\
    - **4 (Mostly): The plan covers the main points of the user input, but might miss a few minor details.\\
    - **3 (Somewhat): The plan addresses some aspects of the user input, but not comprehensively.\\
    - **2 (Slightly): The plan barely touches on the user's input and misses several key points.\\
    - **1 (Not at all): The plan fails to address the user input or is irrelevant.\\

     Ordering (Is the plan well-ordered?)**\\
    - **5 (Definitely):** The ordering does not need any changes.\\
    - **4 (Mostly):** The ordering is generally good, but there might be a few minor adjustments.\\
    - **3 (Somewhat):** I could see reordering some of these, but it would be more of a stylistic change.\\
    - **2 (Slightly):** The ordering could use some improvements, but it's not entirely bad.\\
    - **1 (Not at all):** Ordering is bad or nonsensical.\\

     Completeness (Is the plan complete and informative?)**\\
    - **5 (Definitely):** The plan provides a complete and informative picture of what needs to be done to respond to the user input.\\
    - **4 (Mostly):** The plan is mostly complete and informative, with only a few minor gaps.\\
    - **3 (Somewhat):** The steps are somewhat general, but overall you get what you need. You might need a few minor details.\\
    - **2 (Slightly):** The plan is missing several key details and is not fully clear.\\
    - **1 (Not at all):** The plan is really bland and dominated by unnecessary, irrelevant, and/or repetitive steps, or key steps are missing.\\

     Image Grounded (Can this plan be carried out in the specific setting shown in the image?)**\\
    -- **5 (Definitely): All objects and actions mentioned are clearly present in the image; the plan is specific to the setting seen in the image.\\
    -- **4 (Mostly): The plan makes sense for the setting seen in the image, with only minor mismatches (e.g., one object might be assumed but not shown, or include vague actions to be done in the image presented).\\
    -- **3 (Somewhat): The plan is partially grounded in the setting shown in the image, but some steps rely on questionable assumptions about what’s available or possible to be done.\\
    -- **2 (Slightly): Several actions or objects don’t appear to match the specific setting in the image, making the plan hard to execute as described.\\
    -- **1 (Not at all): The plan feels unrealistic or unrelated to the specific setting in the image—objects are used that are not in the image, actions are implausible or vauge, or it seems like the setting was ignored entirely.\\
    
     Overall (Is the plan overall good?)**\\
    - **5 (Definitely):** The plan is overall good. A good plan should be well-ordered, complete, and contain no repetitive or unnecessary steps.\\
    - **4 (Mostly):** The plan is mostly good. It’s generally well-organized and complete but could use some improvements in detail or clarity.\\
    - **3 (Somewhat):** The steps are somewhat general, but overall you get what you need.\\
    - **2 (Slightly):** The plan is lacking in key details, and some steps feel unnecessary or unclear, but it somewhat meets the requirements.\\
    - **1 (Not at all):** The plan is really bland and not good with repetitive or unnecessary steps.\\
    ---

     **Example 1 Input**\\
    {
        "user\_input": "Can you take my picture with this background?",\\
        "high\_level\_plan": [\\
            "1. Navigate to the Arch",\\
            "2. Position at the Ideal Angle",\\
            "3. Adjust Camera Settings",\\
            "4. Capture the Panoramic Photo"\\
        ]
    }

     **Example 1 Output**\\
        {
        "Coverage": 5, \\
        "Coverage Explanation": "The plan is completely relevant to the user input.", \\
        "Ordering": 5, \\
        "Ordering Explanation": "The steps are in the correct order.", \\
        "Completeness": 4, \\
        "Completeness Explanation": "The plan is mostly complete but lacks specific details about how to adjust the settings.", \\
        "Image Grounded": 4,\\
        "Image Grounded Explanation": "The plan includes only objects in the setting, but it maybe be hard to navigate through the rocks without more directions.",\\
        "Overall": 4,\\
        "Overall Explanation": "The plan is mostly good with minor gaps in detail."\\
    }
   
        **Example 2 Input**\\
    {
        "user\_input": "I'm going on a roadtrip, can you grab me a snack?",\\
        "high\_level\_plan": [\\
            "1. Navigate to the Fridge",\\
            "2. Open the Fridge",\\
            "3. Grab a Grape",\\
        ]
    }

     **Example 2 Output**\\
        {
        "Coverage": 4, \\
        "Coverage Explanation": "Although the plan does get food, one grape might not be enough food for a roadtrip.", 
        "Ordering": 5, \\
        "Ordering Explanation": "The steps are in the correct order.", \\
        "Completeness": 2, \\
        "Completeness Explanation": "The plan does not bring the food to the human.", \\
        "Image Grounded": 5,\\
        "Image Grounded Explanation": "The plan includes objects in the setting.",\\
        "Overall": 3,\\
        "Overall Explanation": "The plan is only slightly address the user input but does not complete it."
    }
    
    Respond strictly in JSON format with the key "Coverage", "Coverage Explanation", "Ordering", "Ordering Explanation", "Completeness", "Completeness Explanation", "Image Grounded", "Image Grounded Explanation", "Overall", and "Overall Explanation". Do not use any markdown formatting or code block symbols (such as triple backticks).
\end{prompt}

\subsection{Embodied Agents Details}\label{appx:exp_details_embodied_agents}
In our second set of experiments we aimed to see how our \ourmethod might result in better downstream performance for embodied agents. We used two simulated experiments to test this hypothesis. 

\begin{figure}[htbp]
    \centering
    \begin{subfigure}[b]{0.23\textwidth}
        \includegraphics[width=\linewidth]{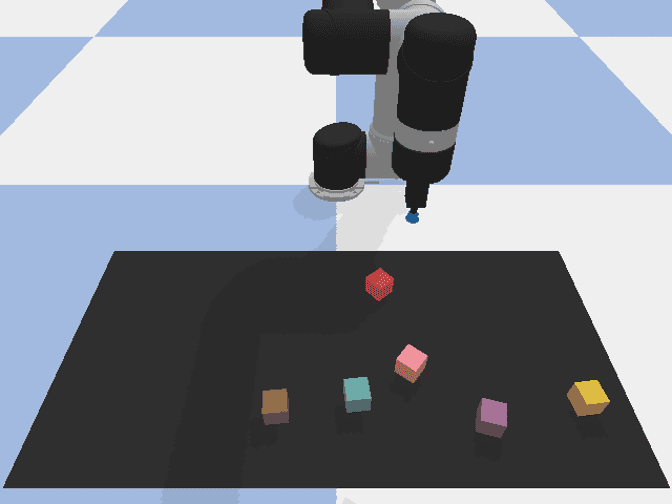}
        \caption{6 Block Image 1}
    \end{subfigure}
    \begin{subfigure}[b]{0.23\textwidth}
        \includegraphics[width=\linewidth]{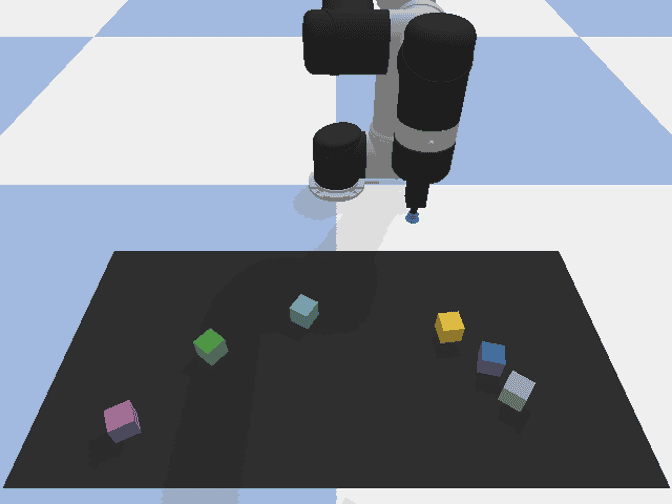}
        \caption{6 Block Image 2}
    \end{subfigure}
    \begin{subfigure}[b]{0.23\textwidth}
        \includegraphics[width=\linewidth]{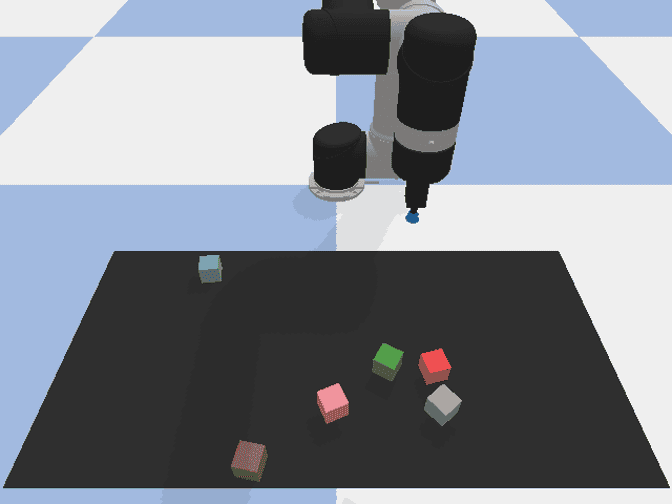}
        \caption{6 Block Image 3}
    \end{subfigure}
    \begin{subfigure}[b]{0.23\textwidth}
        \includegraphics[width=\linewidth]{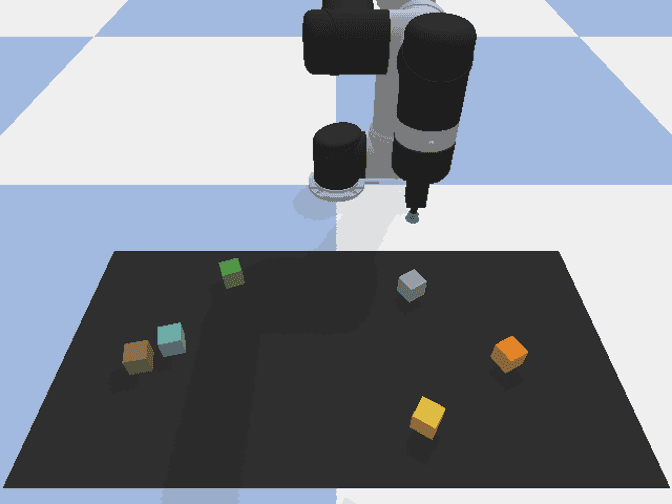}
        \caption{6 Block Image 4}
    \end{subfigure}

    \vspace{1em} %

    \begin{subfigure}[b]{0.23\textwidth}
        \includegraphics[width=\linewidth]{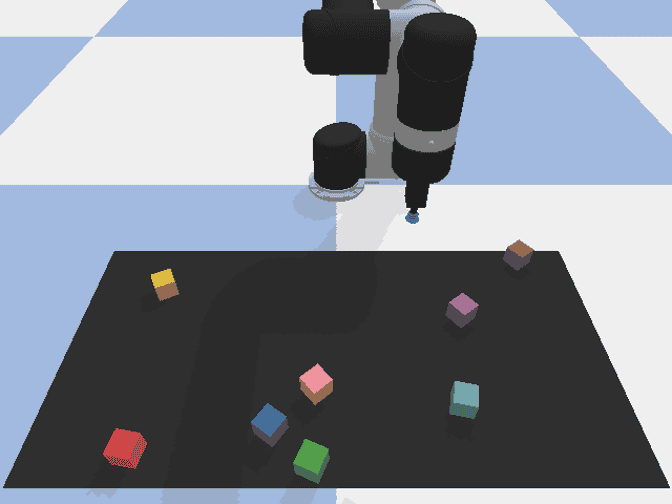}
        \caption{8 Block Image 1}
    \end{subfigure}
    \begin{subfigure}[b]{0.23\textwidth}
        \includegraphics[width=\linewidth]{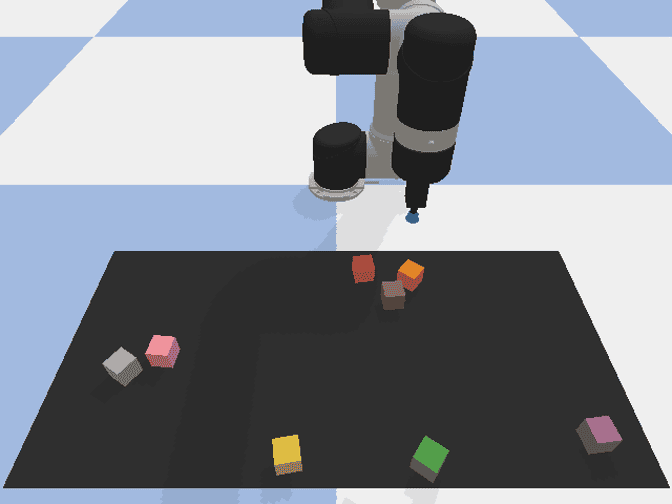}
        \caption{8 Block Image 2}
    \end{subfigure}
    \begin{subfigure}[b]{0.23\textwidth}
        \includegraphics[width=\linewidth]{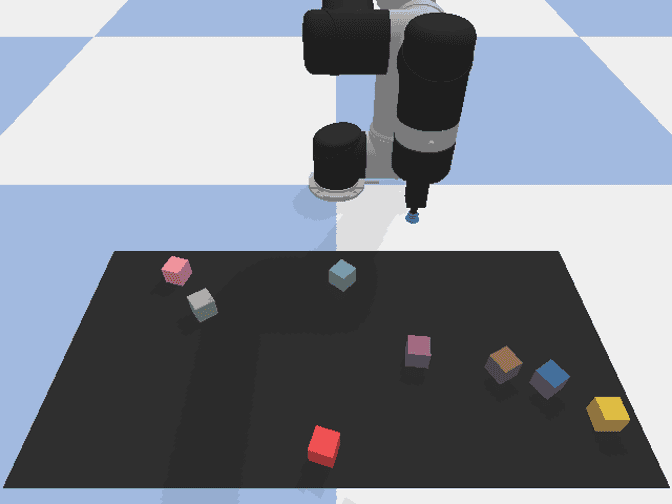}
        \caption{8 Block Image 3}
    \end{subfigure}
    \begin{subfigure}[b]{0.23\textwidth}
        \includegraphics[width=\linewidth]{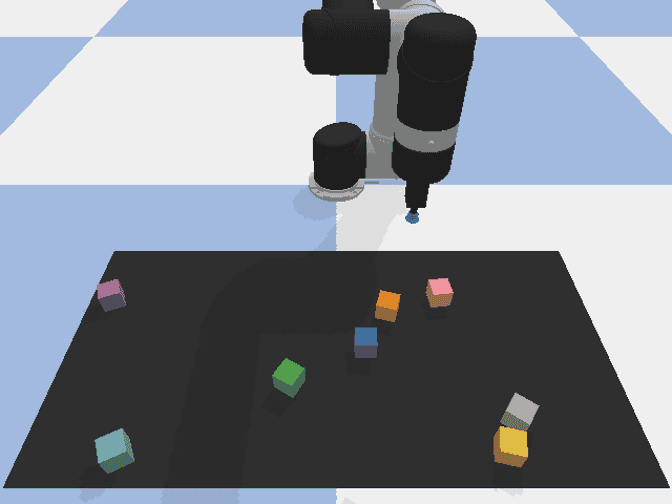}
        \caption{8 Block Image 4}
    \end{subfigure}

    \caption{Eight initial scenes used for the block manipulation task.}
    \label{fig:block_images}
\end{figure}

\paragraph{Evaluation Set}
We used two distinct simulation environments for evaluation: (1) block manipulation tasks from Ravens \cite{zeng2020transporter}, and (2) complex, hierarchical tasks from HAMSTER \cite{li2024hamster}.
For the Ravens environment, we curated 14 unique manipulation goals, each paired with 8 different initial block configurations involving 6 or 8 blocks—yielding a total of $n=112$ samples. Each configuration had blocks of unique colors.  \cref{fig:block_images} shows the 8 individual block scenes and here is the full list of 14 goals are:
\begin{itemize}[noitemsep, topsep=0pt]
    \item Form a shape of an uppercase X with the blocks.
    \item Form a shape of an uppercase O with the blocks.
    \item Form a shape of an uppercase Y with the blocks.
    \item Form a shape of an uppercase V with the blocks.
    \item Form a shape of an uppercase W with the blocks.
    \item Form a diagonal line.
    \item Form two diagonal lines.
    \item Form two vertical lines.
    \item Form two horizontal lines.
    \item Create a smiley face.
    \item Create a frowning face.
    \item Form a shape of triangle with the blocks.
    \item Form the shape of a house.
    \item Form a rainbow.
\end{itemize}. 

\begin{figure*}[htbp]
    \centering
    \begin{subfigure}[t]{0.19\textwidth}
        \includegraphics[width=\linewidth]{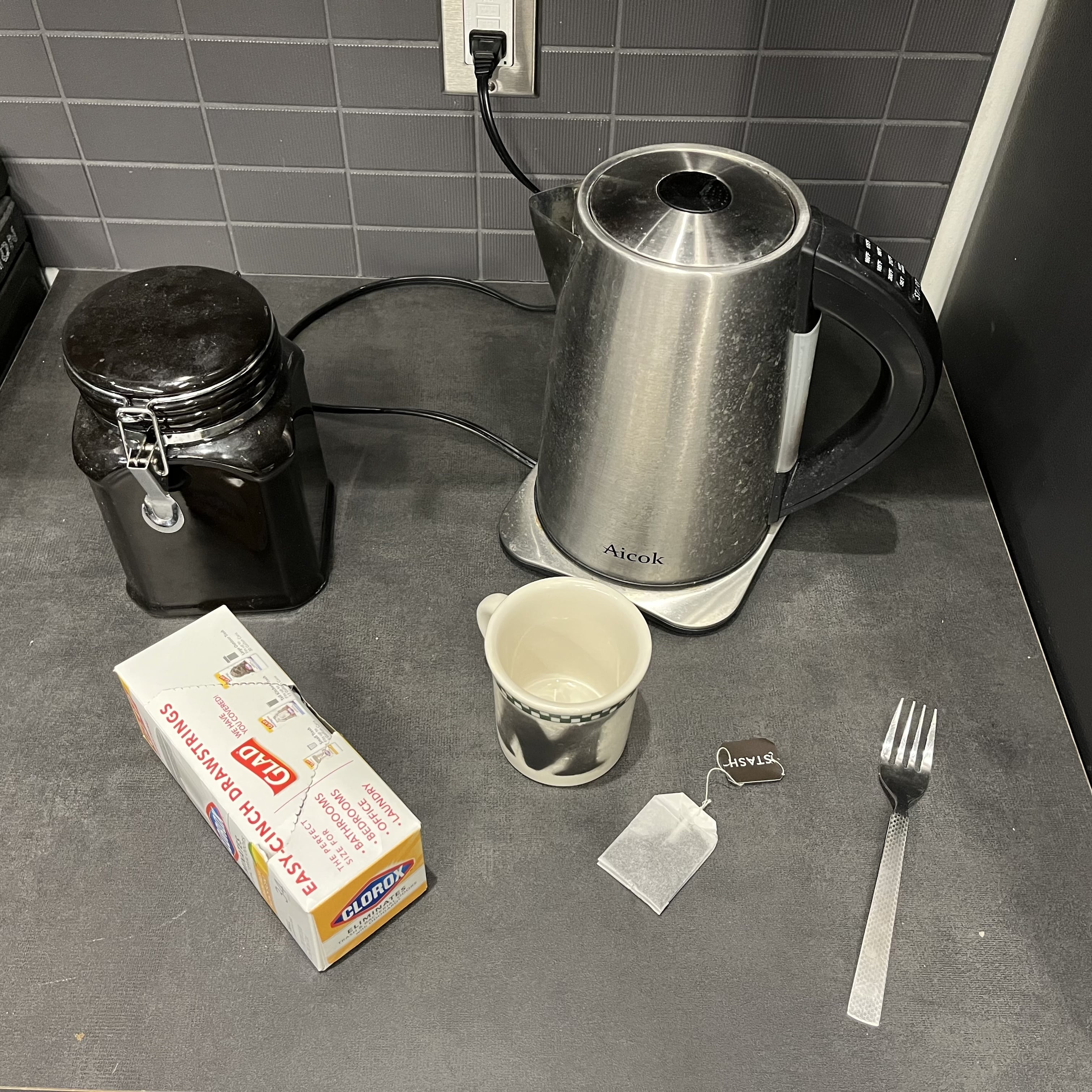}
        \caption{There is water in the kettle. Can you make me a cup of tea?}
    \end{subfigure}
    \begin{subfigure}[t]{0.19\textwidth}
        \includegraphics[width=\linewidth]{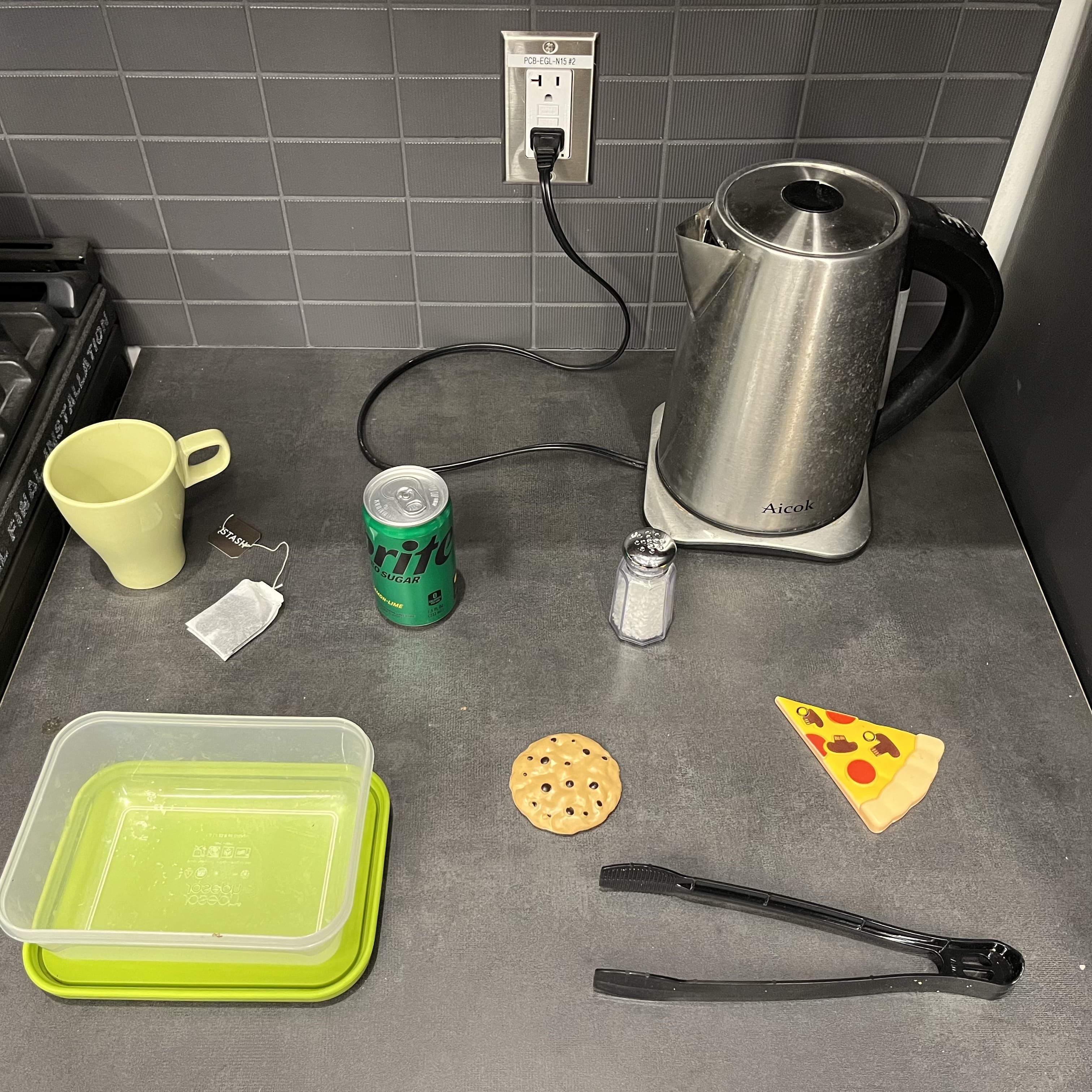}
        \caption{I need to pack a lunch for my kid in the clear tupperware. Only include foods a kid would like.}
    \end{subfigure}
    \begin{subfigure}[t]{0.19\textwidth}
        \includegraphics[width=\linewidth]{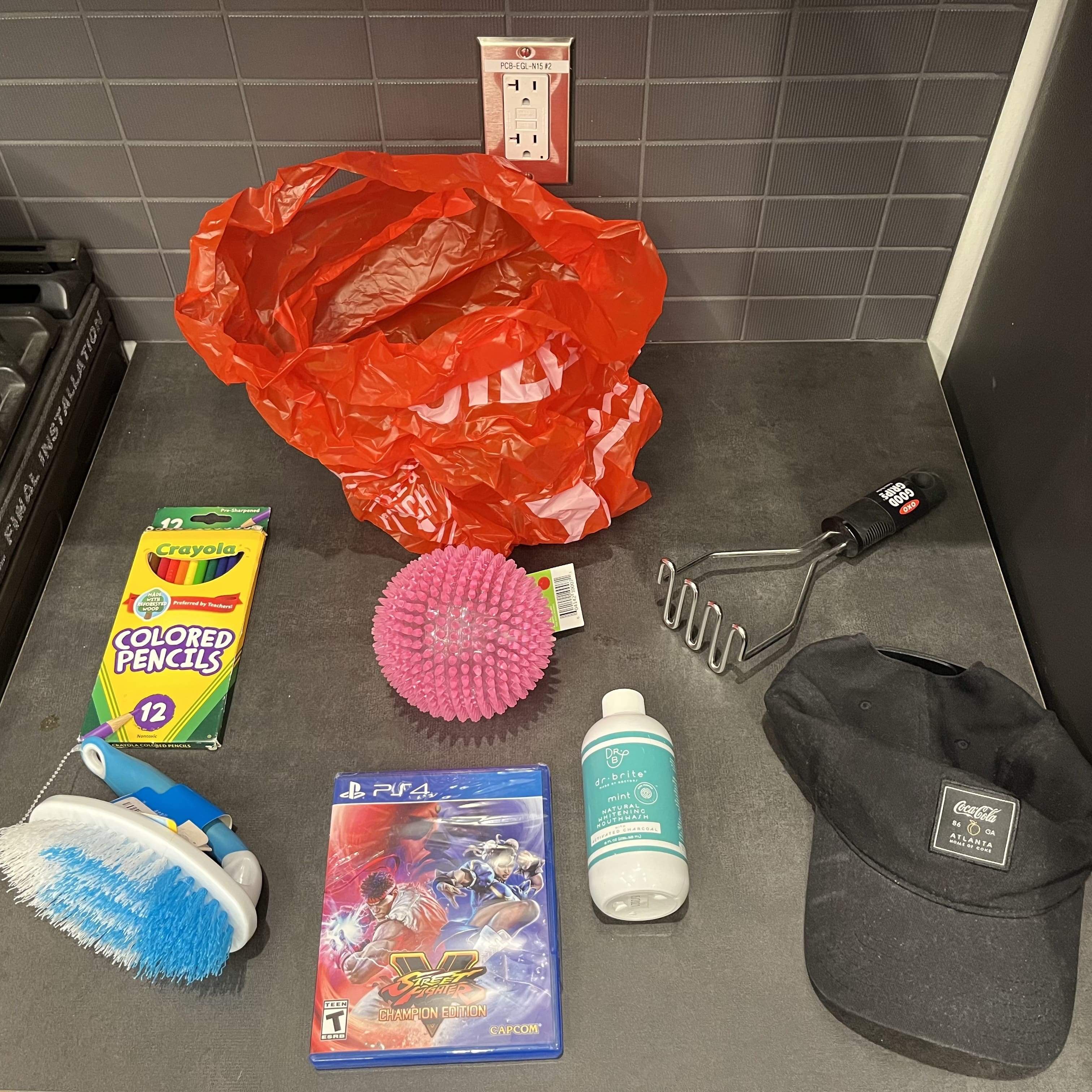}
        \caption{Pack items a kid would like into the red bag. If you do not know what an object is, don't include it. }
    \end{subfigure}
        \begin{subfigure}[t]{0.19\textwidth}
        \includegraphics[width=\linewidth]{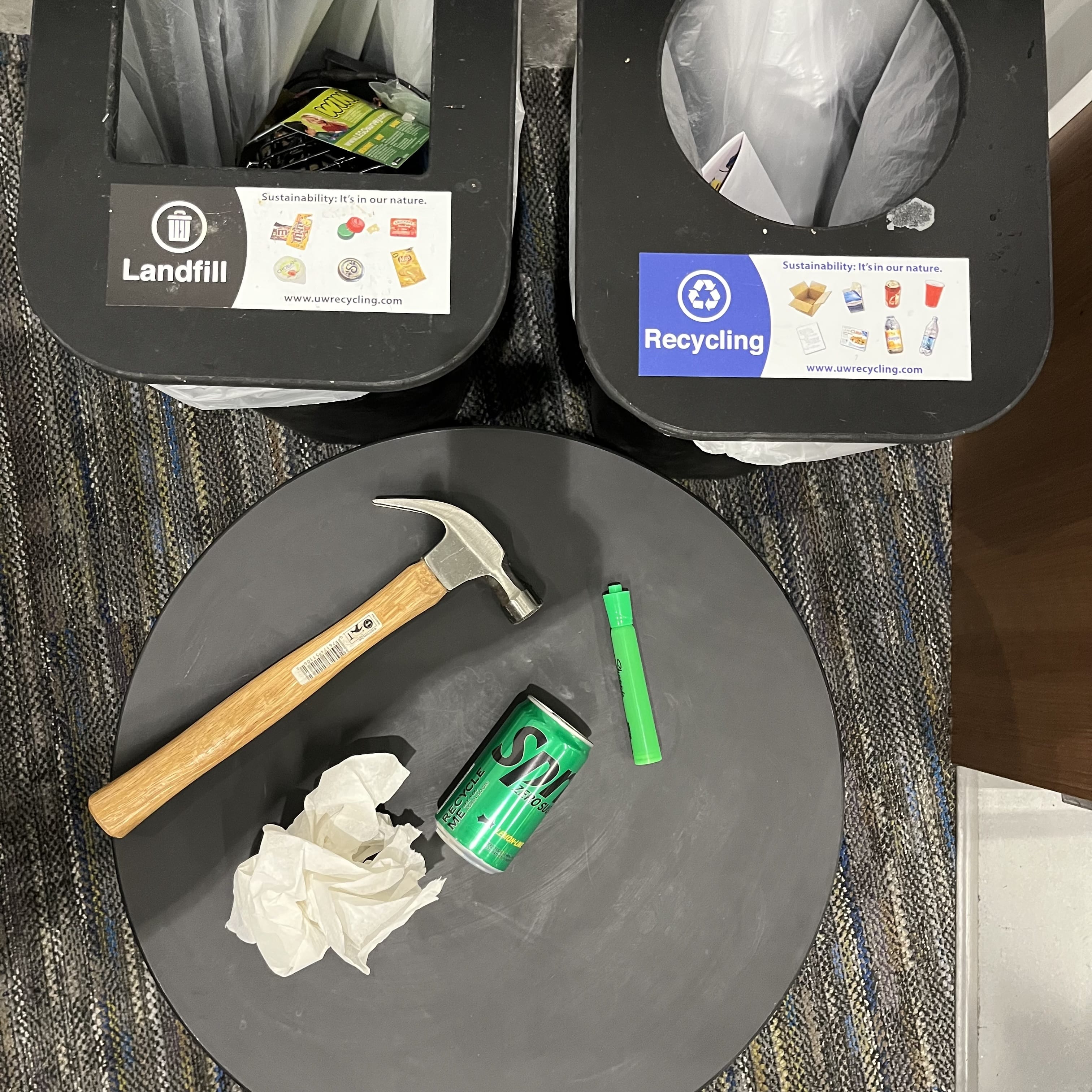}
        \caption{Please organize the trash into the right bins. Note that the paper towel is not used.}
    \end{subfigure}
    \begin{subfigure}[t]{0.19\textwidth}
        \includegraphics[width=\linewidth]{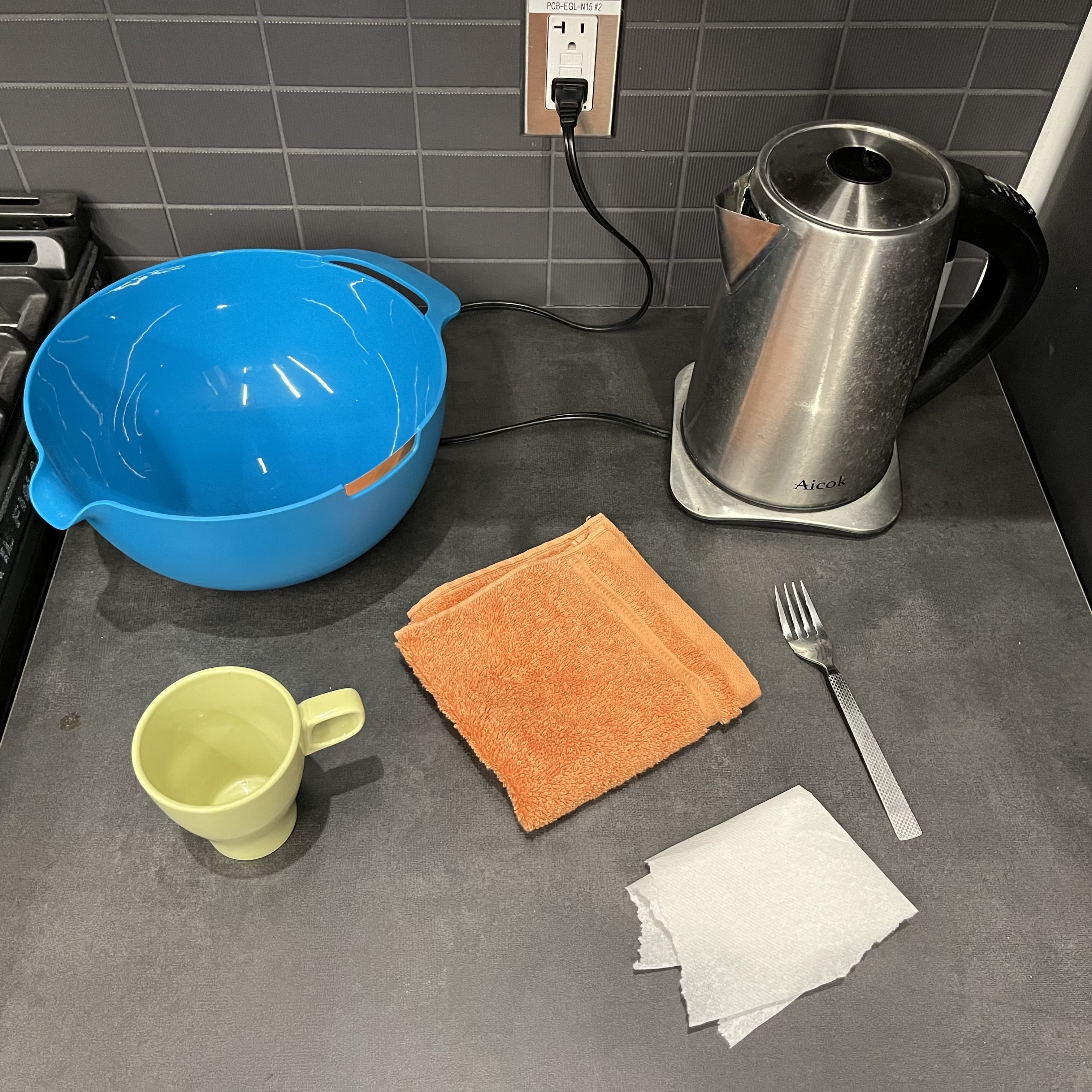}
        \caption{Can you place objects in the bowl in order to clean the counter?}
    \end{subfigure}
    \vspace{1em} %

    \begin{subfigure}[t]{0.19\textwidth}
        \includegraphics[width=\linewidth]{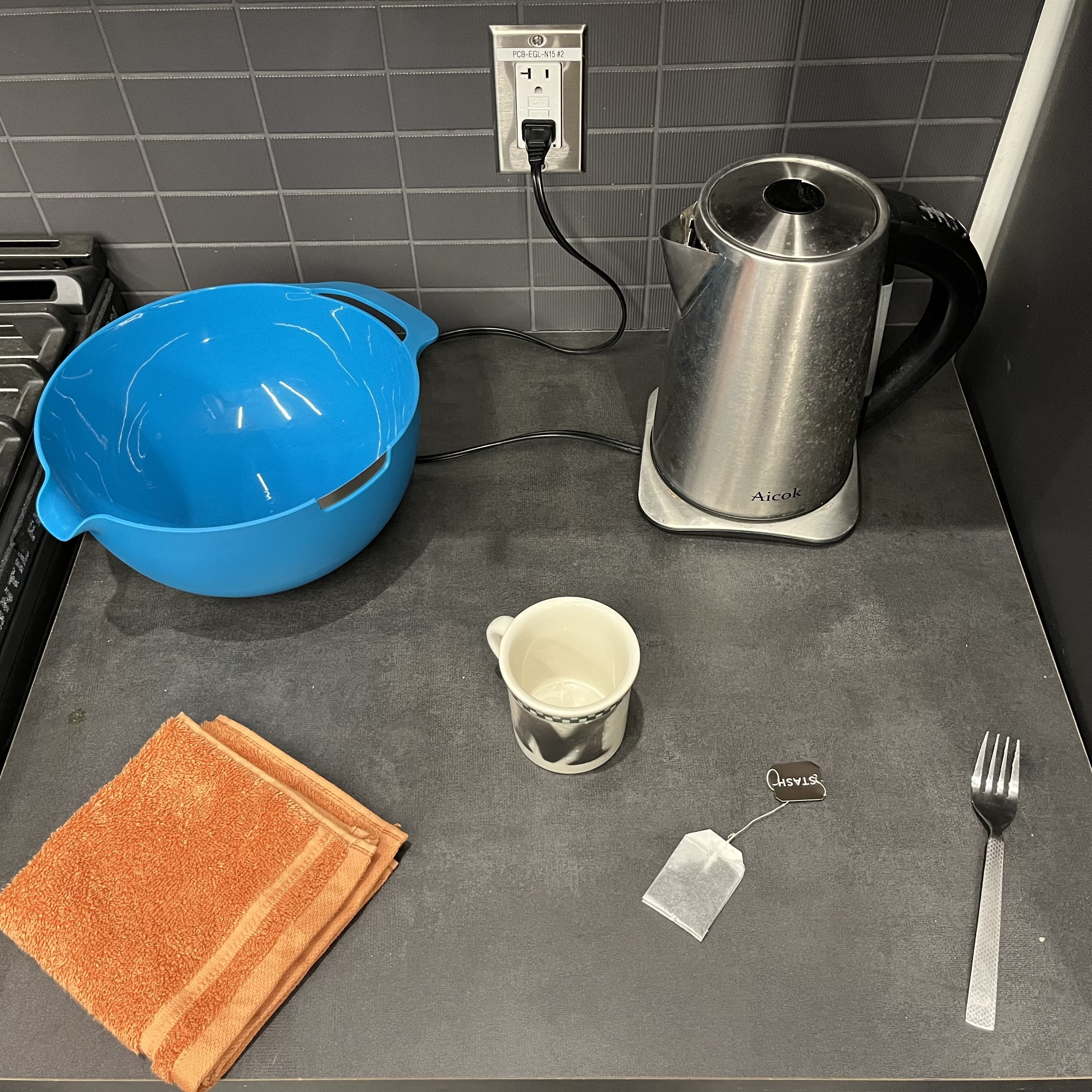}
        \caption{Can you clean the counter to look less messy? No need to wipe the counter, just consolidate all the things. }
    \end{subfigure}
    \begin{subfigure}[t]{0.19\textwidth}
        \includegraphics[width=\linewidth]{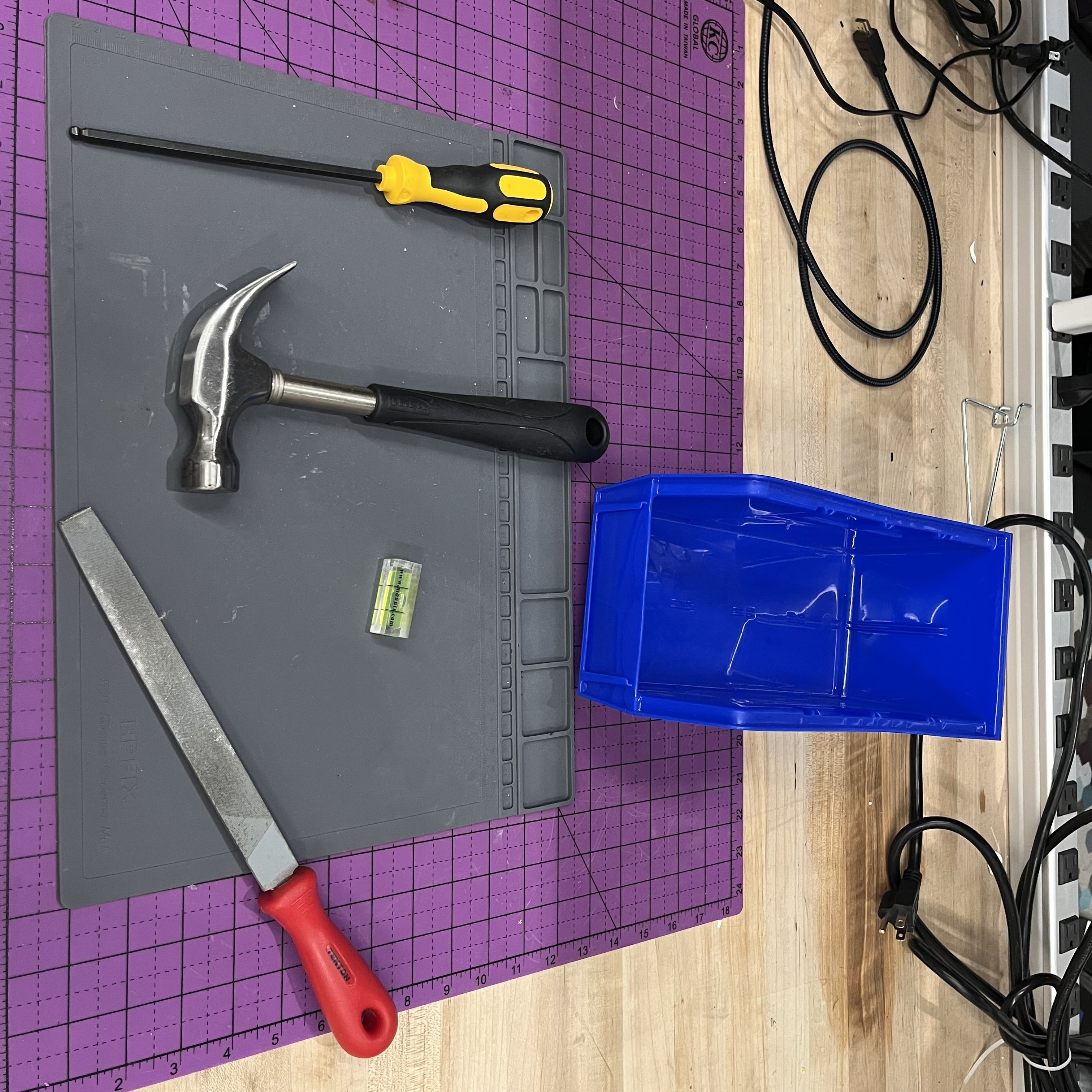}
        \caption{Can you declutter the workshop table?}
    \end{subfigure}
    \begin{subfigure}[t]{0.19\textwidth}
        \includegraphics[width=\linewidth]{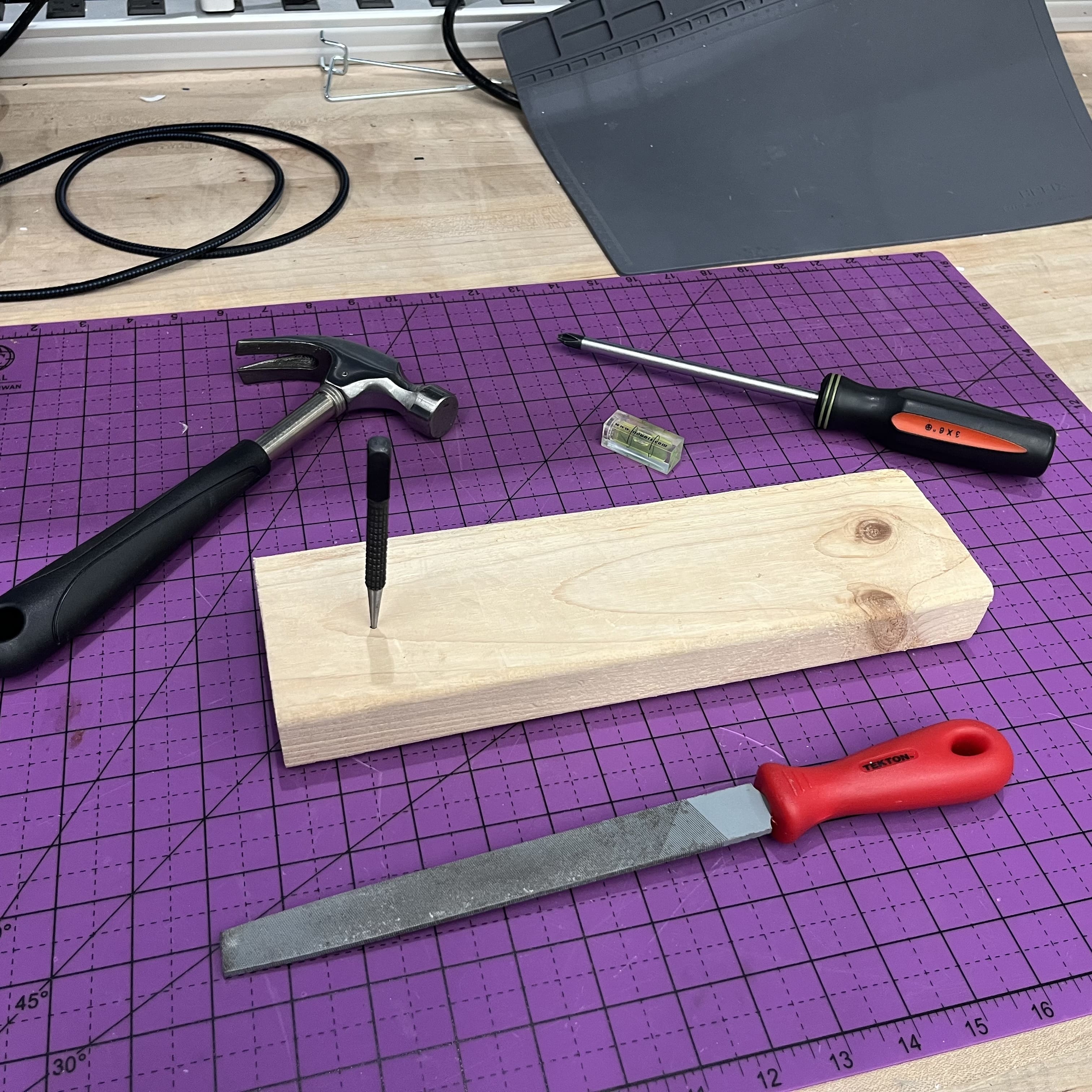}
        \caption{Hammer the nail in the wood.}
    \end{subfigure}
    \begin{subfigure}[t]{0.19\textwidth}
        \includegraphics[width=\linewidth]{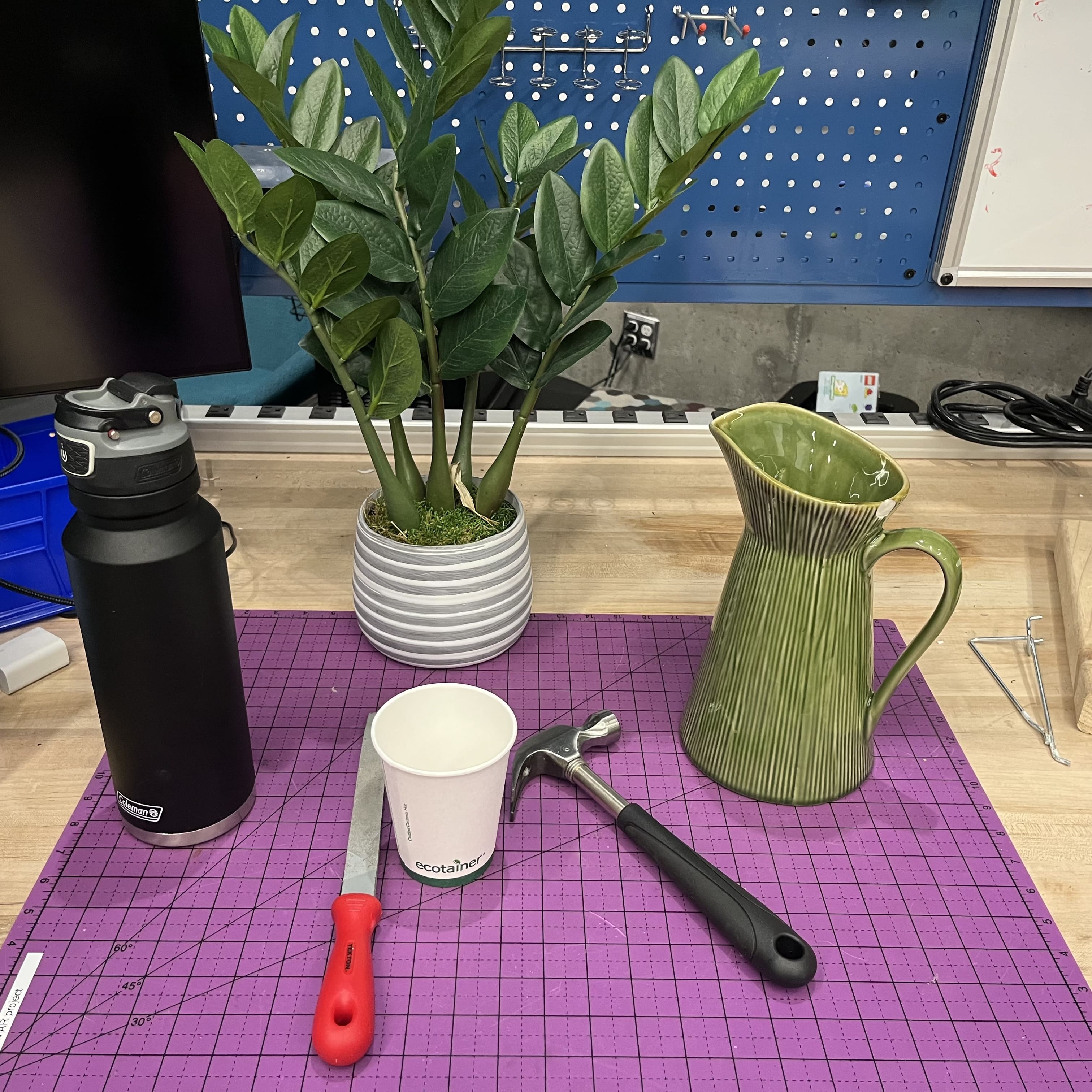}
        \caption{What is the easiest way to water the plant?}
    \end{subfigure}
    \begin{subfigure}[t]{0.19\textwidth}
        \includegraphics[width=\linewidth]{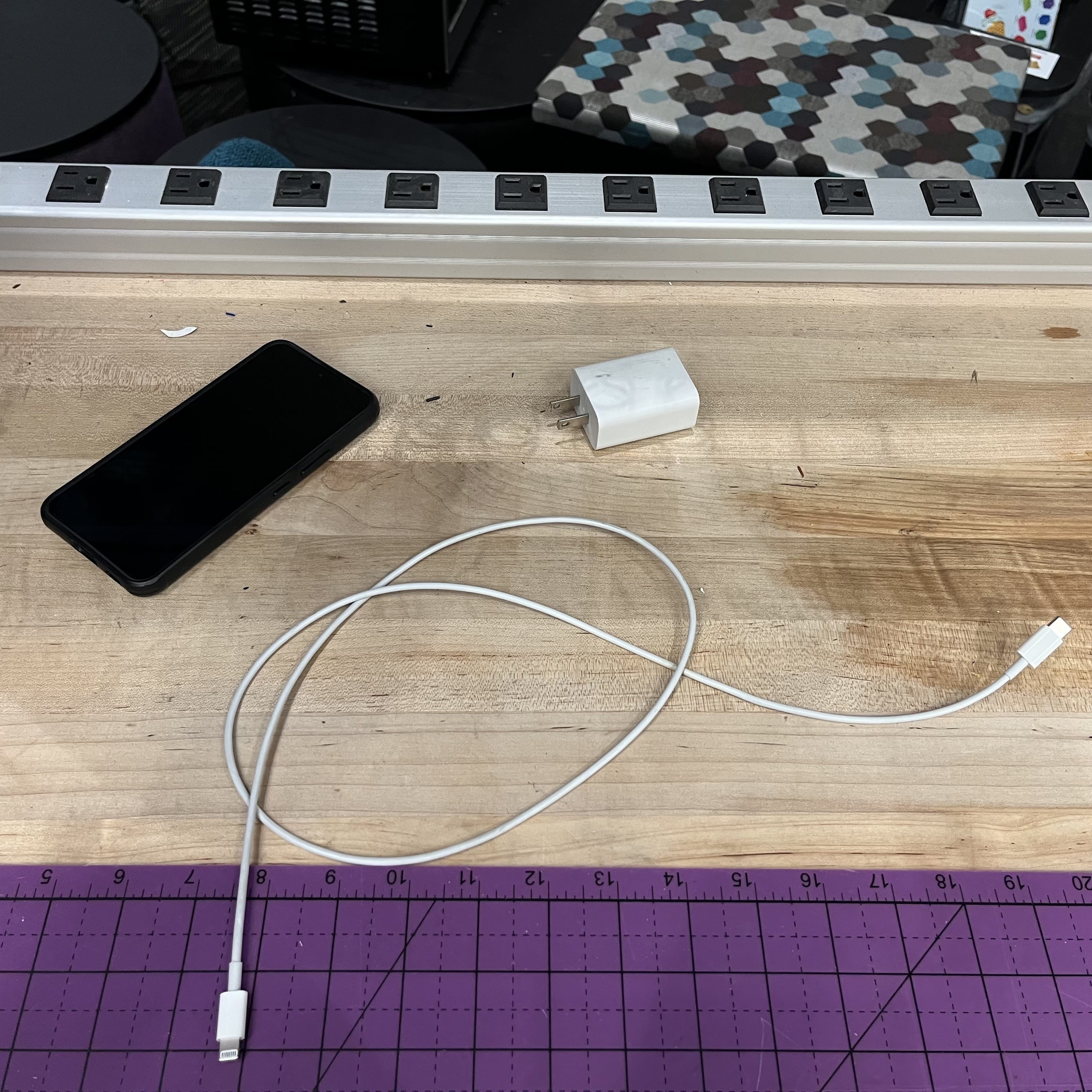}
        \caption{Can you tell me how to charge my phone with what is in this setting?}
    \end{subfigure}
    \label{fig:hamster_images}
    \caption{Images used in the real-world simulation experiments with corresponding goals. }
\end{figure*}

For the hierarchical setting, we designed 10 realistic task scenarios across three environments—kitchen, workshop, and office—each involving a high-level task (e.g., ''Pack items for a children’s lunch''). \cref{fig:hamster_images} shows the 10 realistic task with corresponding goals. 

\paragraph{Metric} For the simulated pick-and-place environment, we run each plan using a code-as-policies simulator \cite{liang2023code} which generated a static image for each step. Then, a human rater evaluated the final configuration, judging whether the plan achieved the stated goal. We then calculated the average number of samples where the code-as-policy successfully ran the plan and achieved the final state. 
For the real-world settings, we used \citet{li2024hamster} to generate a trace path for each step in each generated plan. Then, a human raters assessed whether each individual step was completed successfully by the generated trace. We then indicated the success rate, which is the number of traces that were deemed successful for a step divided by all steps. Note, we did not include steps that would not result in a trace such as "Move to <object>". 
\paragraph{Baselines} We compared the plans generated by SelfReVision with the initial base plan created by the model. For this task we used only Gemma 12B and 27B \cite{gemma}.

\subsection{Software} We used Python 3.12.9, Pytorch 2.6.0, and HuggingFace Transformers 4.51.0.All code is licensed under the
Apache License 2.0.

\subsection{Hardware} All experiments were run on a cluster with 24 NIVIDIA A100 GPUs with 80B memory. For most inference jobs we used one GPU but for 72B models we needed two GPUs. For supervised fine-tuning, we used on GPU for Qwen 3B, two GPUs for Gemma 4B and Qwen 7B, four GPUs for Gemma 12B. The training for four epochs took about two days. 

\subsection{Artifact Terms of Use} 
Places365 \cite{zhou2017places}: MIT License
\section{LLM-as-Judge Analysis}\label{appx:llm_as_judge}
In our study, we used LLM-as-Judge as the main metric for comparison between plans. In this section, we outline our process for evaluating the robustness of using an LLM instead of human raters. 

We did a test on a sample of $n=30$ examples where we evaluated the quality of two robot plans: Plan 0 and Plan n using both human annotators and GPT-4o as an LLM-as-a-Judge. To reduce positional bias during annotation, we randomly assigned these two plans to anonymized labels Plan A and Plan B for each sample shown to human raters. Each plan pair (Plan A and Plan B) is scored on five criteria: Coverage, Ordering, Completeness, Image Groundedness, and Overall. The full annotation instruction can be found in \Fref{appendix:fig:annotation-instruction}.

To measure agreement, we collected annotations from three human annotators and three GPT-4o runs at temperature 0.6. Model outputs were generated using identical prompts and image inputs, with variation arising only from randomized sampling. This setup allowed us to capture inter-model variability due to sampling while maintaining a consistent evaluation protocol.

We chose the Brennan-Prediger coefficient as our agreement metric because it adjusts for chance agreement and handles categorical labels (Plan A", Plan B", or "Tie"). Unlike raw accuracy, it remains robust under label imbalance and is well-suited for comparing multiple raters with potentially different labeling tendencies.

 We report Brennan-Prediger agreement coefficients \citep{brennan1981agreement} between all pairs of raters. The top-level results are summarized below, where we report the mean pairwise agreement between:

\begin{enumerate}[noitemsep, topsep=0pt]
\item Human-Human pairs (3 combinations)
\item Model-Model pairs (3 combinations)
\item Human-Model pairs (9 combinations)
\end{enumerate}

These are computed for each of the five evaluation criteria, and the table below reflects averages across the respective pairings. 

\begin{table*}[h]
\centering
\small
\begin{tabular}{lccc}
\toprule
\textbf{Criterion} & \textbf{Human-Human} & \textbf{Model-Model} & \textbf{Human-Model} \\
\midrule
Coverage        & 0.750 & 0.900 & 0.600 \\
Ordering        & 0.475 & 0.950 & 0.450 \\
Completeness    & 0.425 & 0.850 & 0.558 \\
Image Grounded  & 0.575 & 0.800 & 0.567 \\
Overall         & 0.250 & 0.950 & 0.442 \\
\bottomrule
\end{tabular}
\caption{Brennan-Prediger agreement coefficients for human-human, model-model, and human-model rater pairs, averaged across all combinations and 30 plan comparison samples. GPT-4o was run with temperature 0.6.}
\label{tab:llm_judge_agreement}
\end{table*}
To better understand how often annotators reached full consensus, we measured the percentage of plan pairs where all three human annotators selected the same label:

\begin{itemize}[noitemsep, topsep=0pt]
\item \textbf{Coverage:} 60\% agreement 
\item \textbf{Ordering:} 43\% agreement
\item \textbf{Completeness:} 40\% agreement
\item \textbf{Image Groundedness:} 50\% agreement
\item \textbf{Overall:} 27\% agreement
\end{itemize}

Model-model agreement reflects intra-model consistency under sampling variation. The high model-model agreement across criteria (e.g., 0.95 for Ordering and Overall, 0.90 for Coverage) indicates that GPT-4o produces stable and repeatable judgments across independent runs. Moreover, model-human agreement scores are consistently competitive with human-human agreement—e.g., 0.567 vs.\ 0.575 for Image Groundedness, 0.558 vs.\ 0.425 for Completeness, and 0.600 vs.\ 0.750 for Coverage. These results suggest that GPT-4o is not only internally consistent but also meaningfully aligned with human judgment, supporting its use as a reliable automated judge in comparative plan evaluation tasks.

\begin{figure}
\begin{tcolorbox}

You will be given a user input and two corresponding plans (Plan A and Plan B) with high-level steps that can be used by a robot to respond to the user input in a specific setting. I will also provide an image of the setting when available.

Your task is to evaluate which plan is better based on the following criteria:

\#\#\# Coverage (Does the plan fully address the user input?)
- Does the plan thoroughly address all aspects of the user input without omissions?
- Does the plan cover the main points of the user input, or does it miss details?

\#\#\# Ordering (Is the plan well-ordered?)
- Is the sequence of steps logical and efficient?
- Would any reordering of steps improve the plan?

\#\#\# Completeness (Is the plan complete and informative?)
- Does the plan provide a complete picture of what needs to be done?
- Are the steps specific and detailed enough?
- Are there any gaps in the plan?

\#\#\# Image Groundedness (Can this plan be carried out in the specific setting shown in the image?)**
- Are all objects and actions mentioned clearly present or possible in the given setting in the image?
- Is the plan specific and well grounded to the setting seen in the image?

\#\#\# Overall Assessment
- Considering all criteria above, which plan is better overall?
\end{tcolorbox}
\caption{The instruction given to the human annotators}
\label{appendix:fig:annotation-instruction}
\end{figure} 

\end{document}